\documentclass[12pt,a4paper]{article}
\usepackage[colorlinks,citecolor=blue,urlcolor=blue,bookmarks=false,hypertexnames=true]{hyperref} 
\usepackage[round]{natbib}
\usepackage{authblk}
\usepackage[left=3cm,right=3cm,top=3cm,bottom=3cm]{geometry}
\usepackage[latin1]{inputenc}
\usepackage{amsmath}
\usepackage{amsfonts}
\usepackage{amssymb}
\usepackage{graphicx}
\usepackage{amsthm}
\usepackage{thmtools,blindtext}
\usepackage[dvipsnames]{color}
\usepackage{graphicx}
\usepackage{algorithm}
\usepackage{algpseudocode}

\usepackage{cite}
\usepackage{titlesec}
\usepackage{url}
\usepackage{booktabs}
\usepackage{etoolbox, siunitx}
\usepackage{pbox}
\usepackage{color}
\definecolor{darkpastelgreen}{rgb}{0.01, 0.75, 0.24}
\usepackage{multirow}
\robustify\bfseries
\usepackage{romannum}
\usepackage{array,booktabs,multirow}
\usepackage{etoolbox, siunitx}
\usepackage{pbox}
\robustify\bfseries
\usepackage{tikz}
\usetikzlibrary{shapes,arrows,shadows}
\usepackage{lipsum,adjustbox}
\usepackage{pbox}
\usepackage{varwidth}

\newcommand\eqdef{\mathrel{\stackrel{\makebox[0pt]{\mbox{\normalfont\tiny def}}}{=}}}
\newcommand*{\argmin}{\operatornamewithlimits{argmin}\limits}
\newcommand*{\argmax}{\operatornamewithlimits{argmax}\limits}

\newcommand*{\subjectto}{\operatornamewithlimits{\text{ subject to }}\limits}

\newcommand*{\diag}{\operatornamewithlimits{diag}\limits}

\newcommand\norm[1]{\left\lVert#1\right\rVert}
\newcommand\inner[1]{\left\langle#1\right\rangle}

\newcommand*{\rank}{\operatornamewithlimits{rank}\limits}
\newcommand*{\prox}{\operatornamewithlimits{Prox}\limits}
\newcommand*{\fro}{\operatornamewithlimits{F}\limits}
\newcommand*{\dom}{\operatornamewithlimits{dom}\limits}
\newcommand*{\card}{\operatornamewithlimits{card}\limits}

\newtheorem{theorem}{Theorem}

\title{Convex Latent Effect Logit Model via Sparse and Low-rank Decomposition}
\date{}

\author[1]{Hongyuan Zhan\thanks{Corresponding author. Email: hongyuan.zhan@gmail.com. This work was done while the author was at Penn State University.}}
\author[1]{Kamesh Madduri}
\author[2]{Venkataraman Shankar}
\affil[1]{Facebook Inc.}
\affil[2]{Penn State University, Department of Computer Science and Engineering}
\affil[2]{Texas Tech University, Department of Civil, Environmental, and Construction Engineering}

\begin{document}
\maketitle

\begin{abstract}
In this paper, we propose a convex formulation for learning logistic regression model (logit) with latent heterogeneous effect on sub-population. In transportation, logistic regression and its variants are often interpreted as discrete choice models under utility theory \citep{mcfadden2001economic}. Two prominent applications of logit models in the transportation domain are traffic accident analysis and choice modeling. In these applications, researchers often want to understand and capture the individual variation under the same accident or choice scenario. The mixed effect logistic regression (mixed logit) is a popular model employed by transportation researchers. To estimate the distribution of mixed logit parameters, a non-convex optimization problem with nested high-dimensional integrals needs to be solved. Simulation-based optimization is typically applied to solve the mixed logit parameter estimation problem, i.e., via Monte Carlo simulation of the marginal log-likelihood function. Despite its popularity, the mixed logit approach for learning sub-population or individual heterogeneity has several downsides. First, the parametric form of the distribution requires domain knowledge and assumptions imposed by users, although this issue can be addressed to some extent by using a non-parametric approach. Second, the optimization problems arise from parameter estimation for mixed logit and the non-parametric extensions are non-convex, which leads to unstable model interpretation. Third, the simulation size in simulation-assisted estimation lacks finite-sample theoretical guarantees and is chosen somewhat arbitrarily in practice. To address these issues, we are motivated to develop a formulation that models the latent individual heterogeneity while preserving convexity, and avoids the need for simulation-based approximation. Our setup is based on decomposing the parameters into a sparse homogeneous component in the population and low-rank heterogeneous parts for each individual.
\end{abstract}

\section{Introduction}
\label{sec:intro}
The mixed effect logistic regression model (mixed logit) has great success in transportation econometrics applications. Mixed logit is a generalized linear mixed model for multinomial response \citep{mccullaghGLM83}. In addition, it can also be derived from discrete choice theory as a random utility model \citep{mcfadden2001economic}. It brings advantages over fixed effect multinomial logistic regression (multinomial logit) due to its flexibility to incorporate individual heterogeneity. In mixed logit, the parameters for each observation are random and their distributions are estimated from data. Therefore, conditioned on the same observed variables, two observations may not share the same outcome due to the differences in individual-level parameters. Under random utility theory, the parameter heterogeneity for each observation models taste variations in the population. In addition, when interpreted as a discrete choice model, mixed logit relaxes the Independence of Irrelevant Alternative (IIA) assumption \citep{macfaddenbook,Tra09}. The flexibility to model individual heterogeneity is important in transportation applications. For example in the analysis of traffic accident outcomes, the effect of age is heterogeneous since people of the same age can have diverse health conditions, which affects the degree of resistance to the impact of crash \citep{kim2008age,swedler2012gender,kim2013driver}.  

The flexibility of applying mixed logit model comes with the difficulties to specifying and estimating the model. In the past twenty years, many advances have been made on understanding the mixing distributions used \citep{cherchi2005assessing,fosgerau2006investigating,neuhaus2011effect,mcculloch2011misspecifying}, and relaxing the restrictions on the parametric form of the distribution, which leads to semi-parametric or non-parametric models \citep{greene2003latent,bansal2018minorization}. There are no consensus on the robustness of mixing distributions. Many authors has discussed about the sensitivity of estimates to the specification of mixing distributions \citep{heckman1984method,hensher2003mixed,agresti2004examples,huang2015model}. On the applications of mixed logit to welfare analysis, it has been shown that the choice of mixing distributions has profound impact \citep{cherchi2005assessing,fosgerau2006investigating,huang2015model}. However, there are practical scenarios in which the influence of mixing distribution misspecification is secondary, pointed out in \citet{mcculloch2011misspecifying}. Nevertheless, many recent efforts have been put on making logit models more robust, from both computational and modeling perspective. A line of work is on the usage of non-parametric estimation for mixing distributions \citep{train2008algorithms,de2010bayesian,bansal2018minorization}. A comparison between parametric and non-parametric estimation of the mixing distributions are provided in \citet{bhat2018new}. An early work along this line is the latent class model proposed by \citet{greene2003latent}, in which the heterogeneous parameters are conditioned on the latent class individuals belonged to. The use of mixture-of-normal for approximating the unknown mixing distributions are studied in \citep{fosgerau2009comparison,keane2013comparing}. More recently, the logit-mixture logit (LML) is proposed in \citet{train2016mixed}. LML is a generalization of the latent class logit \citep{greene2003latent} and many other mixture models \citep{train2008algorithms,fox2011simple,fosgerau2013easy}, using functionals such as splines to approximate the kernel of the probability mass function of discrete mixing distribution. Under very general assumptions, any continuous mixing distribution can be approximated by a discrete logit-mixture with sufficiently large support \citep{train2016mixed}. Therefore, non-parametric extensions can be considered as improvements on the robustness with respect to misspecification of mixing distributions.

Mixed logit and the non-parametric variants are typically estimated with maximum simulated likelihood (MSL) estimator \citep{mcfadden1989mom,hajivassiliou1994classical} or the Expectation-Maximization (EM) algorithm \citep{dempster1977maximum}. The expectation operator is nested inside the logarithm in the maximum likelihood estimator. Therefore, simulations are often used to approximate the expectation, leading to the MSL estimator. A large body of work has been done on making the simulation more efficient and numerically more robust approximation of the expectation integrals in practice. Quasi-Monte Carlo (QMC) methods is studied by several authors \citep{train1999halton,bhat2001quasi,bastin2004estimating,munger2012estimation}. It has been shown that QMC reduces the variance of simulation and the number of draws needed in practice. However, it should be noted that the number of draws have to increase with the sample size in theory \citep{hensher2003mixed,train2009discrete}, and there still lacks theoretical guidelines on the sufficient number of draws required on finite samples. Quasi-Newton methods \citep{shanno1970conditioning,fletcher1970new} or Trust-region methods \citep{wright1999numerical,bastin2006application} are then applied to optimize the simulation-based objective function. The EM algorithm for estimating mixed logit model and its variants are studied in \citet{train2008algorithms,pacifico2012fitting,jagabathula2016nonparametric,vij2017random}. \citet{zhanSGDworking} recently studies stochastic gradient method to estimate mixed logit models with Gaussian mixing distribution. \citet{bansal2018minorization} studies the Minorization-Maximization (MM) algorithm, an extension of EM, for fitting the logit-mixture logit model. Note that the maximum likelihood estimators for mixed logit and the non-parametric variants are non-convex, due to the expectation operator nested inside logarithm. Therefore, both the MSL-based estimation or EM-type algorithms are dependent on the starting point. Hence, the model parameters and the estimated mixing distribution can be unstable because multiple local optima exists in the optimization landscape. The effect of initialization on mixed logit parameter estimation was recently studied by \citet{hole2017use}. 

This paper is motivated to develop a model that allows the analysts to incorporate individual heterogeneity while being convex. Preserving convexity is important for obtaining stable estimation. A naive way to allow heterogeneity in the model is by parameterizing each observation independently. This leads to an over-parametrized model. A remedy is to have heterogeneous parameters for each sub-population or cluster. However, the clusters or grouping effect in the population is unobserved. Existent work marginalized the latent variables which produces non-convex objective functions \citep{greene2003latent,train2008algorithms,fosgerau2009comparison,de2010bayesian,keane2013comparing,train2016mixed,bansal2018minorization}. Therefore, we would like to avoid marginalization in the estimation. Mixing distributions can be considered as probabilistic priors on the latent heterogeneity in the population. Alternatively, \textit{structural prior} is a way to encode general knowledge about the structure of data, frequently used in machine learning and inverse problems \citep{kaipio1999inverse,mansinghka2006structured,huang2011learning,jenatton2011structured,vidal2014structured}. In order to avoid marginalization, we exploit structural priors that can be expressed non-probabilistically. To take advantage of these structural information, our proposed model is estimated using a regularized maximum likelihood approach. In particular, we consider the following structure: the homogeneous effect in the population is sparse and the individual heterogeneity is intrinsically low dimension despite the high dimensionality of ambient space. The reason and theoretical justification for imposing this geometric structure will be made clear later. Our work is related to low-rank and sparse matrix decomposition \citep{candes2009exact,candes2011robust}. In many applications, researchers discovered that corrupted or unobserved data can be recovered almost exactly if the data matrix is approximately low-rank \citep{candes2009exact,CT10,candes2011robust,richard2012estimation,soltanolkotabi2014robust,aybat2015admm}. The Netflix challenge \citep{bennett2007netflix} is probably the most well-known application of this technique. Moreover, data from many real problems is found to distributed on intrinsically low dimensional manifolds even if the observations are high dimensional \citep{levina2005maximum,udell2017why}. We develop a fast optimization algorithm to fit the proposed convex latent effect logit model and evaluate the model on traffic accidents extracted from The Statewide Integrated Traffic Records System (SWITRS) \citep{SWITRS}. The computation of one instance of the model on datasets of 10000 samples can be finished in minutes, whereas estimating a mixed logit model on datasets of similar size with the same set of variables takes several hours using commercial software NLOGIT \citep{kim2013driver,nlogit} .

The remaining of this paper is organized as follows: section \ref{sec:background} reviews background knowledge about logit models; section \ref{sec:splrdecomp} introduces our sparse and low-rank decomposition-based convex objective function to incorporate heterogeneity, and explains why low-rankness is a broader geometry description for several different data generating process; section \ref{sec:opt} describes the optimization algorithms for learning the parameters in the model. A greedy scheme for hyperparameter tuning is also introduced in section \ref{sec:opt}; section \ref{sec:case} presents case studies using our proposed model for traffic accident analysis, followed by the conclusion in section \ref{sec:conclusion}. 

\section{Backgrounds}
\label{sec:background}
We provide a brief review for multinomial logit and mixed logit model in this section, and point out the mathematical challenges. In the following sections, let $(\mathbf{x}_n,\mathbf{y}_n)$ a pair of observations, where $\mathbf{x}_n \in \mathbb{R}^{p}$ is the observed variables (features) and $y_n$ is a categorical response from the set $\mathcal{I}=\{1,2,\cdots,I\}$. For example, in choice modeling, $y_n$ represents the decision made by an agent in observation; in traffic crash modeling, $y_n$ denotes the severity category suffered from a crash. The probability of observation response $k \in \mathcal{I}$ conditioned on the variables $\mathbf{x}_n$ and parameter $\boldsymbol{\theta}$ is modeled by the multinomial distribution
\begin{equation}
 P(y_n = k | \mathbf{x}_n, \boldsymbol{\theta}) = \frac{\exp\big( V_k(\mathbf{x}_n,\boldsymbol{\theta}) \big) }{\sum_{j=1}^I \exp\big( V_j(\mathbf{x}_n,\boldsymbol{\theta}) \big)}
\end{equation}  
In discrete choice literature, $\{V_j\}_{j=1}^I$ are usually referred as the propensity functions for making decisions $j \in \mathcal{I}$. The conditional utility \footnote{better interpreted as ``harms" in traffic accidents} from decision $j$ is 
\begin{equation}
U_{j}(\mathbf{x}_n,\boldsymbol{\theta}) = V_j(\mathbf{x}_n,\boldsymbol{\theta}) + \epsilon.
\end{equation}

Variants of logit models differs in how the propensity functions are modeled and the methods used for estimating $\boldsymbol{\theta}$. In multinomial logistic regression (MNL), the propensity functions are parametrized by linear functionals with parameters $\boldsymbol{\theta} := \big(\alpha^{(j)},\boldsymbol{\beta}^{(j)} \big)_{j=1}^I$, 
\begin{equation}
V_j\big(\mathbf{x}_n, \alpha^{(j)},\boldsymbol{\beta}^{(j)}\big) =\alpha^{(j)} + \mathbf{x}^T_n \boldsymbol{\beta}^{(j)}.
\label{eq:MNLpropensity}
\end{equation}
Equation (\ref{eq:MNLpropensity}) implies that two observations have identical probability distribution of outcomes if $\mathbf{x}_n = \mathbf{x}_{n'}$. This is a restrictive assumption that can be violated in the presence of individual heterogeneity. Mixed logit models relax this restriction by placing probability distributions on the parameters, such that individual $n$ has heterogeneous parameters $\boldsymbol{\theta}_n$ governed by a probability distribution $\boldsymbol{\theta}_n \sim f(\boldsymbol{\theta} | \boldsymbol{\psi})$. $f$ is usually referred as the mixing distribution and the parameters $\boldsymbol{\psi}$ for $f$ are to be estimated. Normal, log-normal, and truncated normal distributions are among the popular choices of parametric mixing distributions \citep{Tra09}. On the other hand, $f$ can be constructed via a non-parametric approach and via discrete mixture of distributions \citep{fosgerau2009comparison,fosgerau2013easy,train2016mixed,vij2017random}. Let $N$ denote the sample size. The unobserved heterogeneous parameters $\{\boldsymbol{\theta}_n \}_{n=1}^N$ are usually integrated out, yielding the marginal log-likelihood estimator:
\begin{equation}
 \hat{\boldsymbol{\psi}} \in \argmax_{\boldsymbol{\psi}} \sum_{n=1}^N \log \Big[ \mathbb{E}_{ \boldsymbol{\theta}_n \sim f(\boldsymbol{\theta} | \boldsymbol{\psi}) } \Big( P(y_n | \mathbf{x}_n, \boldsymbol{\theta}_n)   \Big)  \Big].
\label{eq:marginal}
\end{equation} 
There are two main mathematical challenges in solving the optimization problem in (\ref{eq:marginal}). First, $(\ref{eq:marginal})$ is usually non-concave. Second, the expectation taken over the distribution of $\boldsymbol{\theta}_n$ often has no closed-form expressions and requires further approximation. $(\ref{eq:marginal})$ is usually replaced by the Maximum Simulated Likelihood (MSL) estimator \citep{mcfadden1989mom,hajivassiliou1994classical} using (quasi-) Monte Carlo simulation to approximate the expectation. The Expectation Maximization (EM) \citep{dempster1977maximum} iterative procedure is another option.

\section{Sparse and Low-rank Decomposition}
\label{sec:splrdecomp}
We describe our sparse and low-rank decomposition approach for modeling heterogeneity in this section. Let $\boldsymbol{\beta}_n$ denote the parameter for the $n$-th individual. For each observation $n\in\{1,2,\cdots,N\}$, we decompose the individual parameters by
\begin{equation}
\boldsymbol{\beta}_n = \boldsymbol{\mu} + \boldsymbol{\upsilon}_n.
\label{eq:separation}
\end{equation}  
The individual parameter $\boldsymbol{\beta}_n$ is separated into a homogeneous effect vector $\boldsymbol{\mu}$, and a heterogeneous counter-part $\boldsymbol{\upsilon}_n$ for individual $n$. Without any assumptions on $\{\boldsymbol{\upsilon}_n\}_{n=1}^N$, it is impossible to learn the homogeneous-heterogeneous effect decomposition since equation (\ref{eq:separation}) over-parametrizes the data. In mixed logit, the degrees of freedom in $\{\boldsymbol{\beta}_n\}_{n=1}^N$ is constrained by the mixing distributions used. However, this leads to an non-convex objective function as discussed earlier. Instead of imposing mixing distirbutions, we will penalize the degree of freedom of the model by imposing geometry constraints. We first describe the notations before introducing the formulation. Let

\begin{equation}
\mathbf{\Upsilon}:= [\boldsymbol{\upsilon}_1,\boldsymbol{\upsilon}_2,\cdots,\boldsymbol{\upsilon}_N] \in \mathbb{R}^{pI\times N}
\end{equation}
be a matrix appending the heterogeneous parameters $\boldsymbol{\upsilon}_n$ for the individuals as columns. Denote the homogeneous effect for the $j$-th category as $\boldsymbol{\mu}^{(j)} \in \mathbb{R}^p$, and let 
\begin{equation}
\mathbf{U} = [\boldsymbol{\mu}^{(1)},\boldsymbol{\mu}^{(2)},\cdots,\boldsymbol{\mu}^{(I)} ] \in \mathbb{R}^{p\times I} 
\end{equation}
be a matrix collecting the homogeneous effects. Let $\mathbf{U}_i \in \mathbb{R}^I$ denote the $i$-th row of $\mathbf{U}$. The following constrained optimization problem is considered.

\begin{equation}
\begin{aligned}
\min_{\boldsymbol{\alpha},\mathbf{U},\mathbf{\Upsilon}} \frac{1}{N}\sum_{n=1}^N & -\log \Bigg( \frac{\exp{  \big(\alpha^{(t_n)} +{\boldsymbol{\mu}^{(t_n)}}^T \mathbf{x}_n  +{\boldsymbol{\upsilon}^{(t_n)}_n}^T \mathbf{x}_n } \big) }{\sum_{j=1}^I \exp{  \big(\alpha^{(j)} +{\boldsymbol{\mu}^{(j)}}^T \mathbf{x}_n  +{\boldsymbol{\upsilon}^{(j)}_n}^T \mathbf{x}_n } \big)} \Bigg) \\
\subjectto & \sum_{i=1}^p\norm{\mathbf{U}_i}_{2}  \leq \tau_1\\ 
& \rank(\mathbf{\Upsilon})  \leq \tau_2.
\end{aligned}
\label{eq:lowrank_constraint}
\end{equation} 
where $\tau_1\geq 0$, and $\tau_2\geq 0$ are hyper-parameters to be selected. $\sum_{i=1}^p\norm{\mathbf{U}_i}_{2}$ is the so-called group-$\ell_1$ norm used by $\textit{group Lasso}$ \citep{yuan2006model,friedman2010note}. When $\tau_1$ is smaller than minimum group-$\ell_1$ norm of the solutions to the unconstrained problem, the optima $\mathbf{U}^*$ in (\ref{eq:lowrank_constraint}) forces $\mathbf{U}^*_i$ into a vector of zeros for some $i\in\{1,2,\cdots,p\}$. Therefore, the optimal solution $\mathbf{U}$ is sparse and there are no homogeneous effect for variable $i$ if $\mathbf{U}^*_i= \mathbf{0}$. When $\tau_2 \ll \min(pI,N)$, the heterogeneity effects $[\boldsymbol{\upsilon}_n]_{n=1}^N$ are constrained in a low-rank vector space. In the next section, we explain why imposing low-rank structural assumptions helps uncovering the separation between homogeneous and heterogeneous effect. Problem (\ref{eq:lowrank_constraint}) is non-convex due to the rank constraint. A convex relaxation to (\ref{eq:lowrank_constraint}) will be presented in section \ref{sec:opt}.

\subsection{Why Low-Rankness?} 
\label{sec:Latent_whylowrank}
In this section, we explain situations where low-rank heterogeneous effect arises.
\subsubsection{Gaussian mixing with approximately low-rank covariance}
The homogeneous-heterogeneous effect decomposition in (\ref{eq:separation}) resembles the linear representation for Gaussian mixing variables. In Gaussian mixed logit model, the random parameters can be decomposed as
\begin{equation}
\boldsymbol{\beta}_n = \boldsymbol{\mu} + \boldsymbol{\eta}_n, \text{ where } \boldsymbol{\eta}_n \sim \mathcal{N}(\mathbf{0},\mathbf{I}).
\end{equation}
Hence the mean of the Gaussian distributions corresponds to the homogeneous effect, and 
\begin{equation}
\mathbf{\Upsilon} = \mathbf{\Gamma} \mathbf{H}
\end{equation}
where $\mathbf{H}:=[\boldsymbol{\eta}_1,\boldsymbol{\eta}_1,\cdots,\boldsymbol{\eta}_N]\in \mathbb{R}^{pI\times N}$. Low-rank matrix of mixing effects $\mathbf{\Upsilon}$ occurs when the covariance matrix is (approximately) low-rank. This happens when the covariance matrix is a superposition of a low-rank component and a sparse components \citep{chandrasekaran2012latent,richard2012estimation,meng2014learning,oymak2015simultaneously,chen2015exact}, or when the covariance matrix has a block diagonal structure where the variables within each block are highly correlated and variables across different blocks are independent \citep{saunderson2012diagonal,liutkus2017diagonal}.  

\subsubsection{Latent clustered heterogeneity}
Suppose the population forms clusters, such that the predictive variables have identical or similar influence on individuals belonging to the same cluster. Let $\{\mathcal{C}_l\}_{p=1}^{\tau}$ be a partition of the data, i.e., $\mathcal{C}_{l} \cap \mathcal{C}_{} = \emptyset$ if $l\neq l'$, $\cup_{l=1}^{\tau} |\mathcal{C}_l| = N$. If the individual heterogeneities are identical on each cluster, then

\begin{equation}
\boldsymbol{\upsilon}_i = \boldsymbol{\upsilon}_j \hspace{0.5cm} \forall~  i,j \in C_l, ~l=1,\cdots, \tau.
\label{eq:clusterlatent}
\end{equation}
Therefore, observations in the same cluster have identical columns in $\mathbf{\Upsilon}$. As a result, $\rank(\mathbf{\Upsilon}) \leq \tau$. (\ref{eq:clusterlatent}) can be relaxed slightly, consider
\begin{equation}
\norm{\boldsymbol{\upsilon}_i - \boldsymbol{\upsilon}_j}_{\ell_p} \leq \epsilon \hspace{0.5cm} \forall~  i,j \in C_l, ~l=1,\cdots, \tau.
\label{eq:approxclusterlatent}
\end{equation} 
(\ref{eq:approxclusterlatent}) leads to grouping of the heterogeneous effects, and forces the latent effects in the same group are stays close. $\mathbf{\Upsilon}$ can be approximated by a low-rank matrix in this case. Note that latent clustering of heterogeneity is also an implicit assumption in discrete mixture multinomial logits \citep{greene2003latent,train2008algorithms,pacifico2012fitting,jagabathula2016nonparametric,vij2017random}.

\subsubsection{Latent matrix factorization}
Suppose $\rank(\mathbf{\Upsilon}) = \tau$ with $\tau \leq \min(pI,N)$. $\mathbf{\Upsilon}$ can be equivalently expressed as 
\begin{equation}
\mathbf{\Upsilon} = \mathbf{W}\mathbf{V}^T,	
\end{equation} 
where $\mathbf{W} \in \mathbb{R}^{pI\times \tau}$, $\mathbf{V} \in \mathbb{R}^{N\times \tau}$ and both $\mathbf{W}$ and $\mathbf{V}$ has full column-rank. The matrix $\mathbf{V}$ is usually interpreted as latent loadings, and $\mathbf{W}$ are scores for each of the basis in $\mathbf{V}$. This is sometimes referred as a matrix factorization representation \citep{lee1999learning,lee2001algorithms,ding2005equivalence,ding2008equivalence,abdi2010principal,gillis2014and}. Under this interpretation, individual heterogeneities are created through different linear combinations of the same latent sources. 

\subsection{Convex relaxation}
Optimization with matrix rank constraints is NP-Hard. Finding the exact solutions to problem (\ref{eq:lowrank_constraint}) is difficult and can be unpractical. Therefore, we use a convex program to approximate the original formulation. Let $\norm{A}_*$ be the \textit{nuclear norm} for matrix $A$ define as following.
\begin{equation}
\norm{A}_* = \sum_{i=1}^{\tau} \sigma_i,
\end{equation}
where $\tau$ is the rank of $A$ and $\sigma_i$ is the $i$-th largest singular value of $A$. The nuclear norm $\norm{A}_*$ is a convex relaxation for the rank function $\rank(A)$ \citep{fazel2001rank,recht2008necessary,recht2010guaranteed} and has been successfully applied on many problems involving the rank function as objective or constraints \citep{candes2009exact,candes2011robust,soltanolkotabi2014robust,aybat2014efficient,udell2017why,chandrasekaran2012latent,richard2012estimation,meng2014learning,oymak2015simultaneously,chen2015exact} (and references therein). Therefore, we solve the following convex program
\begin{equation}
\min_{\boldsymbol{\alpha},\mathbf{U},\mathbf{\Upsilon}} \frac{1}{N}\sum_{n=1}^N  -\log \Bigg( \frac{\exp{  \big(\alpha^{(t_n)} +{\boldsymbol{\mu}^{(t_n)}}^T \mathbf{x}_n  +{\boldsymbol{\upsilon}^{(t_n)}_n}^T \mathbf{x}_n } \big) }{\sum_{j=1}^I \exp{  \big(\alpha^{(j)} +{\boldsymbol{\mu}^{(j)}}^T \mathbf{x}_n  +{\boldsymbol{\upsilon}^{(j)}_n}^T \mathbf{x}_n } \big)} \Bigg) + \lambda_1 \sum_{i=1}^p\norm{\mathbf{U}_i}_{2} + \lambda_2 \norm{\mathbf{\Upsilon}}_* .
\label{eq:latent_lowrank_sparse}
\end{equation}
(\ref{eq:latent_lowrank_sparse}) reformulates (\ref{eq:lowrank_constraint}) by first transforming the original problem into the Lagrangian form, then applying the nuclear norm relaxation to $\rank{(\mathbf{\Upsilon})}$. (\ref{eq:latent_lowrank_sparse}) is the addition of three convex functions and therefore the convexity is preserved. $\lambda_1, \lambda_2 \in \mathbb{R}_+$ are non-negative hyperparameters correspond to the upper bound $\tau_1$ and $\tau_2$ in the hard-constraint version. 

\section{Optimization}
\label{sec:opt}

\subsection{Proximal Gradient algorithm}
We now describe an efficient algorithm to solve the convex problem (\ref{eq:latent_lowrank_sparse}). Note that the terms $\sum_{i=1}^p\norm{\mathbf{U}_i}_{2}$ and $\norm{\mathbf{\Upsilon}}_*$ are non-smooth. Therefore methods requiring second order continuity, for example quasi-Newton methods \citep{fletcher1970new,broyden1970convergence,shanno1970conditioning} and Trust-region methods \citep{sorensen1982newton,bastin2006application}, cannot be applied. The main steps in the algorithm are to deal with these non-smooth terms.
Problem $(\ref{eq:latent_lowrank_sparse})$ has the form
\begin{equation}
\min_{\boldsymbol{\theta}} \{F(\boldsymbol{\theta})  \equiv f(\boldsymbol{\theta}) + h(\boldsymbol{\theta})\},
\label{eq:composite}
\end{equation} 
where $f$ is a smooth convex part corresponds to the multinomial loss, $h$ is convex but non-smooth part consists of $\big(\lambda_1 \sum_{i=1}^p\norm{\mathbf{U}_i}_{2} + \lambda_2 \norm{\mathbf{\Upsilon}}_*\big)$. Proximal Gradient method (PG) \citep{beck2009fast,cai2010singular,toh2010accelerated,ma2011fixed,jenatton2011proximal} is an efficient first-order algorithm to handle the non-smoothness in $h$. It performs the following simple iteration
\begin{equation}
\boldsymbol{\theta}_{t+1} = {\prox}_{s_t,h}\big(\boldsymbol{\theta}_t - s_t \nabla f(\boldsymbol{\theta}_t)\big)
\label{eq:proxgrad}
\end{equation}
where $s_t$ is the step-size and $\prox(\cdot)$ is the proximal operator associated with $s_t$ and $h(\cdot)$ defined below.
\begin{equation}
{\prox}_{s_t,h}(\boldsymbol{\theta}) = \argmin_{\boldsymbol{\omega}} \Big\{\frac{1}{2}\norm{\boldsymbol{\omega} - \boldsymbol{\theta}}^2_2 + s_t h(\boldsymbol{\theta})\Big\}.
\label{eq:proxoperatpr}
\end{equation}

Suppose $f$ is a continuously differentiable function with Lipschitz gradients, i.e.,
\begin{equation}
\norm{\nabla f{(\boldsymbol{\theta}_1)} - \nabla f{(\boldsymbol{\theta}_2})} \leq L \norm{\boldsymbol{\theta}_1 - \boldsymbol{\theta}_2}
\label{eq:Lipschitz}
\end{equation}
for all $\boldsymbol{\theta}_1, \boldsymbol{\theta}_2 \in \dom{(f)}$. The convergence rate of Proximal Gradient algorithm is given by the theorem below.
\begin{theorem}
\citep{beck2009fast} Suppose $h$ is a convex continuous function and possibly non-smooth, $f$ is a convex continuously differentiable function satisfying the Lipschitz condition in (\ref{eq:Lipschitz}) with Lipschitz constant $L$.  Let $\{\boldsymbol{\theta}_{t}\}_{t\geq 1}$ be a sequence of iterates generated from (\ref{eq:proxgrad}) using constant step-size $s_t=\frac{1}{L}$, and let $\boldsymbol{\theta}^*$ be an optimal solution of (\ref{eq:composite}). Then
$$ F(\boldsymbol{\theta}_t) - F(\boldsymbol{\theta}^*) \leq \frac{L\norm{\boldsymbol{\theta}_0 - \boldsymbol{\theta}^*}}{2t} $$
for any $t\geq 1$.
\label{eq:ISTArate}
\end{theorem}
In particular, Theorem $\ref{eq:ISTArate}$ suggests the Proximal Gradient algorithm requires $\mathcal{O}\left(\frac{1}{\epsilon} \right)$ iterations to achieve $ F(\boldsymbol{\theta}_t) - F(\boldsymbol{\theta}^*) \leq \epsilon$. When the Lipschitz constant is unknown, \citet{beck2009fast} shown a line-search algorithm to select the step-sizes $\{s_t\}_{\geq 1}$, which requires computing the solution to the proximal operator (\ref{eq:proxoperatpr}) for each line-search trial step. Solving the minimization problem in the proximal operation can be costly in some cases. Thus we derive the closed-form solutions for the proximal sub-problems and describe an adaptive step-size heuristic later. The Proximal Gradient method closely resembles the iterations in Gradient Descent, especially when the proximal operator admits close-form solutions. We now apply PG to the convex latent logit model (\ref{eq:latent_lowrank_sparse}). Let $(\boldsymbol{\alpha}_t,\mathbf{U}_t, \mathbf{\Upsilon}_t)$ denote the iterates after the $t$-th PG iteration (\ref{eq:proxgrad}). Applying (\ref{eq:proxgrad}) on (\ref{eq:latent_lowrank_sparse}) leads to the following update steps using intermediate variables $(\widehat{\boldsymbol{\alpha}}_{t},\widehat{\mathbf{U}}_{t}, \widehat{\mathbf{\Upsilon}}_{t})$

\begin{subequations}
	\begin{align}
		(\widehat{\boldsymbol{\alpha}}_{t},\widehat{\mathbf{U}}_{t}, \widehat{\mathbf{\Upsilon}}_{t}) =& (\boldsymbol{\alpha}_t,\mathbf{U}_t, \mathbf{\Upsilon}_t) -  \frac{s_t}{N}\sum_{n=1}^N \nabla \ell \Big(\mathbf{x}_n,t_n; \boldsymbol{\alpha}_t,\mathbf{U}_t, \mathbf{\Upsilon}_t\Big)  \\
		\boldsymbol{\alpha}_{t+1} =& \widehat{\boldsymbol{\alpha}}_{t} \label{eq:alphaupdate} \\ 
		(\mathbf{U}_{t+1}, \mathbf{\Upsilon}_{t+1}) =& \argmin_{\mathbf{U},\mathbf{\Upsilon}} \Big\{ \frac{1}{2}  \Big(\norm{\widehat{\mathbf{U}}_t-\mathbf{U} }^2_{\fro} + \norm{\widehat{\mathbf{\Upsilon}}_t-\mathbf{\Upsilon} }^2_{\fro} \Big) + s_t \Big( \lambda_1 \sum_{i=1}^p\norm{\mathbf{U}_i}_{2} + \lambda_2 \norm{\mathbf{\Upsilon}}_*  \Big)  \Big\} \label{eq:proxi_join}
	\end{align}
\end{subequations}

The sub-problem from the proximal operator (\ref{eq:proxi_join}) can be decoupled into two separate minimization problems:
\begin{subequations}
	\label{eq:proxi_separate}
	\begin{align}
	\mathbf{U}_{t+1} = &\argmin_{\mathbf{U}} \Big\{ \frac{1}{2}  \norm{\widehat{\mathbf{U}}_t-\mathbf{U} }^2_{\fro}  + s_t \lambda_1 \sum_{i=1}^p\norm{\mathbf{U}_i}_{2}    \Big\}
	\label{eq:proxi_U} \\
	\mathbf{\Upsilon}_{t+1} = &\argmin_{\mathbf{\Upsilon}} \Big\{ \frac{1}{2}  \norm{\widehat{\mathbf{\Upsilon}}_t-\mathbf{\Upsilon} }^2_{\fro} + s_t \lambda_2 \norm{\mathbf{\Upsilon}}_*  \Big\}
	\label{eq:proxi_V}
	\end{align}
\end{subequations}
Moreover, both (\ref{eq:proxi_U}) and (\ref{eq:proxi_V}) has closed-form solutions. The optimal solution to (\ref{eq:proxi_U}) can be obtained by solving the proximal operator independently over each row $\{\mathbf{U}_i\}_{i=1}^p$. The sub-problems with respect to $\mathbf{U}_i$ are essentially the proximal operation arisen from the group-Lasso penalty \citep{yuan2006model,jenatton2011proximal}. For each $i=1,2,\cdots,p$, the analytical solution for the proximal operator associated with step-size $s_t$ and $\lambda_1\norm{\cdot}_2$ is
\begin{equation}
{\prox}_{s_t,  \lambda_1\norm{\cdot}_2} \big( \mathbf{U}_{i} \big) = \big( 1 - \frac{s_t \lambda_1}{ \norm{\mathbf{U}_{i}} }  \big)_+ \mathbf{U}_{i},
\label{eq:groupl1thresholding}
\end{equation}  
where the operator $(\cdot)_+$ denotes thresholding $a_+ = \max(0,a)$. The closed-form solution for the proximal operator associated with nuclear norm penalty $\lambda_2\norm{\cdot}_*$ and step-size $s_t$ can be obtained via a generalization of the soft-thresholding procedure in (\ref{eq:groupl1thresholding}). Suppose the Singular Value Decomposition (SVD) for $\boldsymbol{\Upsilon}$ is 
\begin{equation}
\mathbf{\Upsilon} = \mathbf{P}\mathbf{\Sigma}\mathbf{V}^T,
\end{equation}
where $\mathbf{\Sigma}= \diag(\{\sigma_i\}_{i=1}^{\tau})$ is the diagonal matrix of singular values of $\mathbf{\Upsilon}$ and $\rank{(\mathbf{\Upsilon})} = \tau$. The closed-form solution to (\ref{eq:proxi_V}) is given by the Singular-Value Thresholding (SVT) operator \citep{cai2010singular,toh2010accelerated,ma2011fixed},
\begin{equation}
{\prox}_{s_t, \lambda_2 \norm{\cdot}_*}(\mathbf{\Upsilon}) = \mathbf{P} \mathcal{T}_{s_t \lambda_2}(\mathbf{\Sigma})\mathbf{V}^T, \hspace{0.1cm} \text{where }
\mathcal{T}_{s_t \lambda_2}(\mathbf{\Sigma}) = \diag{\{(\sigma_i - s_t\lambda_2)_+\}_{i=1}^{\tau}}.
\label{eq:SVT}
\end{equation}
(\ref{eq:groupl1thresholding}) and (\ref{eq:SVT}) together provide the solutions to the proximal sub-problem (\ref{eq:proxi_join}). 

\subsection{Acceleration, adaptive step-size and randomized SVT}
\label{sec:practicalOPT}
We apply several simple techniques to improve the convergence speed of Proximal Gradient algorithm described in the previous section. Nesterov's momentum method is a well-known technique to derive optimal convergence rate for first-order optimization methods on smooth convex problems \citep{nesterov1983}. The extension to accelerate the convergence on non-smooth objectives via proximal operators are developed by Beck and Teboulle \citep{beck2009fast}, known as the Fast Iterative Shrinkage-Thresholding Algorithm (FISTA). Under constant step-size $s_t = \frac{1}{L}$ or line-searched step-sizes, FISTA achieves improved worse-case convergence rate of $\mathcal{O}\big(\frac{1}{t^2}\big)$. The acceleration technique is applied in our implementation. In practice, the Lipschitz constant might be unknown or hard to compute. On the other hand, backtracking line-search requires computing the proximal operator over a range of different step-sizes. Since the intermediate variable $\widehat{\mathbf{\Upsilon}}_{t}$ changes due to different trial step-size $s_t$ during the line search process, the Singular Value Decomposition of $\widehat{\mathbf{\Upsilon}}_{t}$ needs to be re-computed in each trial step in order to obtain ${\prox}_{s_t, \lambda_2 \norm{\cdot}_*}(\widehat{\mathbf{\Upsilon}}_{t})$. Therefore backtracking line-search becomes costly due to multiple computations of SVD in each iteration. Hence we use an adaptive step-size scheme as a surrogate to line-search. After each iteration, the objective value of (\ref{eq:latent_lowrank_sparse}) is measured, the step-size is halved if the objective value increases compared to the previous iteration. Note that Accelerated Proximal Gradient (APG) is not a strictly descent method, as oppose to Gradient Descent, therefore tentative increments of the function value do occur. This phenomenon is referred as the Nesterov ripples in the literature \citep{nesterov1983,o2015adaptive}. Further, we employ a function value-based restarting criterion introduced in \citep{o2015adaptive}. The algorithm is restarted with the current iterates as re-initialization when ripples are detected, and the step-size is halved. The acceleration momentums are reset at each restart. A partial SVD is needed in order to obtain the proximal solution for the nuclear norm of the heterogeneous effect matrix $\boldsymbol{\Upsilon}$, as shown in eqn. (\ref{eq:SVT}). Note that the dimension of $\boldsymbol{\Upsilon}$ is $pI \times N$. Therefore, SVD computation becomes a bottleneck during the optimization when $pI$ or $N$ is large. To scale up the estimation problem (\ref{eq:latent_lowrank_sparse}) on large datasets, we deploy the randomized Singular Value Decomposition algorithm \citet{halko2011finding} to perform the SVT operation. To obtain an approximate SVD for a matrix $\boldsymbol{A} \in \mathbb{R}^{m\times n}$, we seek for a matrix $\mathbf{Q}$ with orthonormal columns such that 
\begin{equation}
	\norm{\mathbf{A}-\mathbf{Q}\mathbf{Q}^T \mathbf{A} } \leq \epsilon,
\end{equation} 
and $\mathbf{Q}$ approximately spans the range of $\mathbf{A}$. In addition, we would like the computation of $\mathbf{Q}$ to be lightweight, by the mean of random projection. Let $\boldsymbol{\Omega}=[\boldsymbol{\omega}_i]_{i=1}^k \in \mathbb{R}^{n\times k}$ be $k$ standard Gaussian random vectors, where $k=\rank(\mathbf{A})$. Then 
\begin{equation}
	\boldsymbol{Y}=\mathbf{A} \boldsymbol{\Omega}
\end{equation}
forms a matrix with linearly-independent columns with high probability. This idea is formalized in the seminal paper by \citet{halko2011finding}. The randomized SVD algorithm is given in Algorithm \ref{alg:randomizedSVD}. Essentially, Algorithm \ref{alg:randomizedSVD} can be viewed as first performing dimensionality reduction via random projection (line 1-4), then running the deterministic SVD for a matrix of smaller dimension (line 5) and lastly recovering the SVD for the original matrix (line 6-7). Algorithm \ref{alg:SVT} increases the guess for minimum number of singular values required to reach below the threshold and applies randomized SVD until the threshold is found. The Fast Accelerated Proximal Gradient with Adaptive Restart (FAPGAR) algorithm combining these features is listed in Algorithm \ref{alg:FAPG}.  



\begin{algorithm}[!t]
	\caption{Randomized SVD}
	\label{alg:randomizedSVD}
	\textbf{Input}: input matrix $\mathbf{A} \in \mathbb{R}^{m \times n}$, number $k$ of top singular vectors desired\\
	\textbf{Output}: approximate rank-$k$ SVD of $\mathbf{A}$
	\begin{algorithmic}[1]
		\State generate random matrix $\boldsymbol{\Omega} \in \mathbb{R}^{n\times k}$ with i.i.d. standard normal entries.
		\State construct $\mathbf{Y} =  \mathbf{A} \boldsymbol{\Omega}$  \Comment{{\color{blue}random projection}}
		\State QR decomposition $[\mathbf{Q},\mathbf{R}] = \texttt{qr}(\mathbf{Y})$ to obtain orthonormal basis for $\mathbf{Y}$.
		\State construct $\mathbf{B}= \mathbf{Q}^T \mathbf{A}$ \Comment{{\color{blue}smaller dimension than $\mathbf{A}$}}
		\State rank-$k$ SVD for $\mathbf{B}$, $[\mathbf{W}, \mathbf{S}, \mathbf{V}] = \texttt{svd}(\mathbf{B},k)$
		\State $\mathbf{U} = \mathbf{Q} \mathbf{W}$.
		\State output $\mathbf{U},\mathbf{S},\mathbf{V}$.
	\end{algorithmic}
\end{algorithm}

\begin{algorithm}[!ht]
	\caption{Randomized SVT}
	\label{alg:SVT}
	\textbf{Input}: input matrix $\mathbf{A} \in \mathbb{R}^{m \times n}$, threshold $\rho \in \mathbb{R}_+$, initial guess $k$.\\
	\textbf{Output}: approximate Singular-Value Thresholding operation on $\mathbf{A}$
	\begin{algorithmic}[1]
		\State $[\mathbf{U},\mathbf{S},\mathbf{V}] = \texttt{RandomizedSVD}(\mathbf{A},k)$
		\While{$\min(\mathbf{S}) > \rho$}
		\State $k \leftarrow 1.5(k+1)$
		\State $[\mathbf{U},\mathbf{S},\mathbf{V}] = \texttt{RandomizedSVD}(\mathbf{A},k)$		
		\EndWhile
		\State $\mathbf{S} = \diag{\{(\mathbf{S}_{ii} - \rho)_+\}_{i=1}^{\tau}}$
		\State output $\mathbf{U}\mathbf{S}\mathbf{V}^T$
	\end{algorithmic}
\end{algorithm}

\begin{algorithm}[!t]
	\caption{Fast Accelerated Proximal Gradient with Adaptive Restart}
	\label{alg:FAPG}
	\textbf{Input}: number of iterations $T$, initial step-size $s_0$, initialization $\boldsymbol{\alpha}_{0}, \mathbf{U}_{0}, \mathbf{\Upsilon}_{0}$, tolerance $\epsilon_{tol}$, tuning parameters $\lambda_1,\lambda_2 \in \mathbb{R}_+$\\
	\textbf{Output}: solution to (\ref{eq:latent_lowrank_sparse}).
	\begin{algorithmic}[1]
		\State $q_1 = 1$
		\State $\widetilde{\boldsymbol{\alpha}}_{1} = \boldsymbol{\alpha}_{0}$
		\State $\widetilde{\mathbf{U}}_{1} = \mathbf{U}_{0}$
		\State $\widetilde{\mathbf{\Upsilon}}_{1} = \mathbf{\Upsilon}_{0}$
		
		\For{$t$ $=$ $1$ to $T$}
		\State  $(\boldsymbol{\alpha}_{t},\widehat{\mathbf{U}}_{t}, \widehat{\mathbf{\Upsilon}}_{t}) = (\boldsymbol{\alpha}_t,\mathbf{U}_t, \mathbf{\Upsilon}_t) -  \frac{s_t}{N}\sum_{n=1}^N \nabla \ell \Big(\mathbf{x}_n,t_n; \widetilde{\boldsymbol{\alpha}}_t,\widetilde{\mathbf{U}}_t, \widetilde{\mathbf{\Upsilon}}_t\Big)$
		\For{$i=1$ to $p$} \Comment{proximal operator for group $\ell_1$ penalty}
		\State $(\mathbf{U}_i)_{t} = \big( 1 - \frac{s_t \lambda_1}{ \norm{(\widehat{\mathbf{U}}_{i})_t } }  \big)_+ (\widehat{\mathbf{U}}_{i})_t$  
		\EndFor
		
		\State $\mathbf{\Upsilon}_t = \texttt{RandomizedSVT}(\widehat{\mathbf{\Upsilon}}_{t},s_t \lambda_2, \rank{(\mathbf{\Upsilon}_{t-1})+1})$ \Comment{proximal operator for nuclear norm penalty}
		
		\If{$F(\boldsymbol{\alpha}_{t},\mathbf{U}_{t},\mathbf{\Upsilon}_{t}) > F(\boldsymbol{\alpha}_{t-1},\mathbf{U}_{t-1},\mathbf{\Upsilon}_{t-1}) $}
		\State $q_t=1$  \Comment{restart}
		\State $\widetilde{\boldsymbol{\alpha}}_{t+1} = \boldsymbol{\alpha}_{t}$
		\State $\widetilde{\mathbf{U}}_{t+1} = \mathbf{U}_{t}$
		\State $\widetilde{\mathbf{\Upsilon}}_{t+1} = \mathbf{\Upsilon}_{t}$
		\State $s_{t+1} = s_t/2$ \Comment{adjust step-size}
		\Else
		\State $q_{t+1} = \frac{1+\sqrt{1+4 q^2_t}}{2}$ \Comment{acceleration}
		\State $\widetilde{\boldsymbol{\alpha}}_{t+1} = \boldsymbol{\alpha}_{t} + \big( \frac{q_t - 1}{q_{t+1}} \big) \big( \boldsymbol{\alpha}_{t} - \boldsymbol{\alpha}_{t-1}\big)$  
		\State $\widetilde{\mathbf{U}}_{t+1} = \mathbf{U}_{t} + \big( \frac{q_t - 1}{q_{t+1}} \big) \big( \mathbf{U}_{t} - \mathbf{U}_{t-1}\big)$
		\State $\widetilde{\mathbf{\Upsilon}}_{t+1} = \mathbf{\Upsilon}_{t} + \big( \frac{q_t - 1}{q_{t+1}} \big) \big( \mathbf{\Upsilon}_{t} - \mathbf{\Upsilon}_{t-1}\big)$
		\State $s_{t+1} = s_t$
		\EndIf
		\If{$\norm{ \boldsymbol{\alpha}_{t} - \boldsymbol{\alpha}_{t-1}}_2 + \norm{\mathbf{U}_{t} - \mathbf{U}_{t-1}}_{\fro} + \norm{\mathbf{\Upsilon}_{t} - \mathbf{\Upsilon}_{t-1}}_{\fro} < \epsilon_{tol}$}
		\State break
		\EndIf
		\EndFor
	\end{algorithmic}
\end{algorithm}

\section{Greedy Local Continuation for Pathwise Solutions}
Algorithm \ref{alg:FAPG} solves one instance of problem (\ref{eq:latent_lowrank_sparse}), given $\lambda_1$ and $\lambda_2$. Hence, we solve (\ref{eq:latent_lowrank_sparse}) over a range of $\lambda_1$ and $\lambda_2$ using Algorithm \ref{alg:FAPG}, then choose the regularization constants by some goodness-of-fit criteria. Suppose the search space in each $\lambda$ is discretized into $c$ points. Grid search requires calling Algorithm \ref{alg:FAPG} $c^2$ times in order to obtain the solution over the two dimensional grids. This can be prohibitive. In this section, we describe a fast strategy via greedy local search and continuation method to establish the solution over a range of tuning constants.

\subsection{Prediction for new observations}
Given a new unseen sample $\mathbf{x}_n \in \mathbb{R}^p$, the latent heterogeneous effect $\boldsymbol{\upsilon}_n$ needs to be decided. As show in section \ref{sec:Latent_whylowrank}, clustering is one of the situations causing low-rank latent effect. Therefore, given a new observation $\mathbf{x}_n$, we query its $k$-nearest neighbors using distance defined by one minus the cosine similarity $w_{ij}=\frac{\inner{\mathbf{x}_i, \mathbf{x}_{j}}}{\norm{\mathbf{x}_i}\norm{\mathbf{x}_{j}}}$. The heterogeneous effect for the new test sample is set to the weighted average of heterogeneous effect of observations belonged to the neighborhood $\mathcal{N}$ of the new test sample, i.e., $\boldsymbol{\upsilon}_n = \sum_{j \in \mathcal{N} } w_j \boldsymbol{\upsilon}_j / \sum_{j \in \mathcal{N}} w_j $.

\subsection{Greedy local continuation}
For Lasso or other $\ell_1$ type penalized fixed effect generalized linear models, Friedman et al. \citep{friedman2007pathwise,friedman2010regularization}. showed that $\norm{\beta(\lambda)}_1$ forms a continuous path of the regularization constant $\lambda$. Therefore, the model parameters stay close when change in the penalty constant is small enough. Based on this observation, Friedman et al. proposed warm start strategy to compute the solution over the entire path of regularization hyperparameters, where the optimal solution from a neighboring $\lambda$ is used as initialization \citep{friedman2007pathwise,friedman2010regularization}. This idea is extended to the nuclear norm penalized problem in \citep{ma2011fixed}.    

\begin{figure}[ht!]
	\centering
	\includegraphics[width=0.5\textwidth]{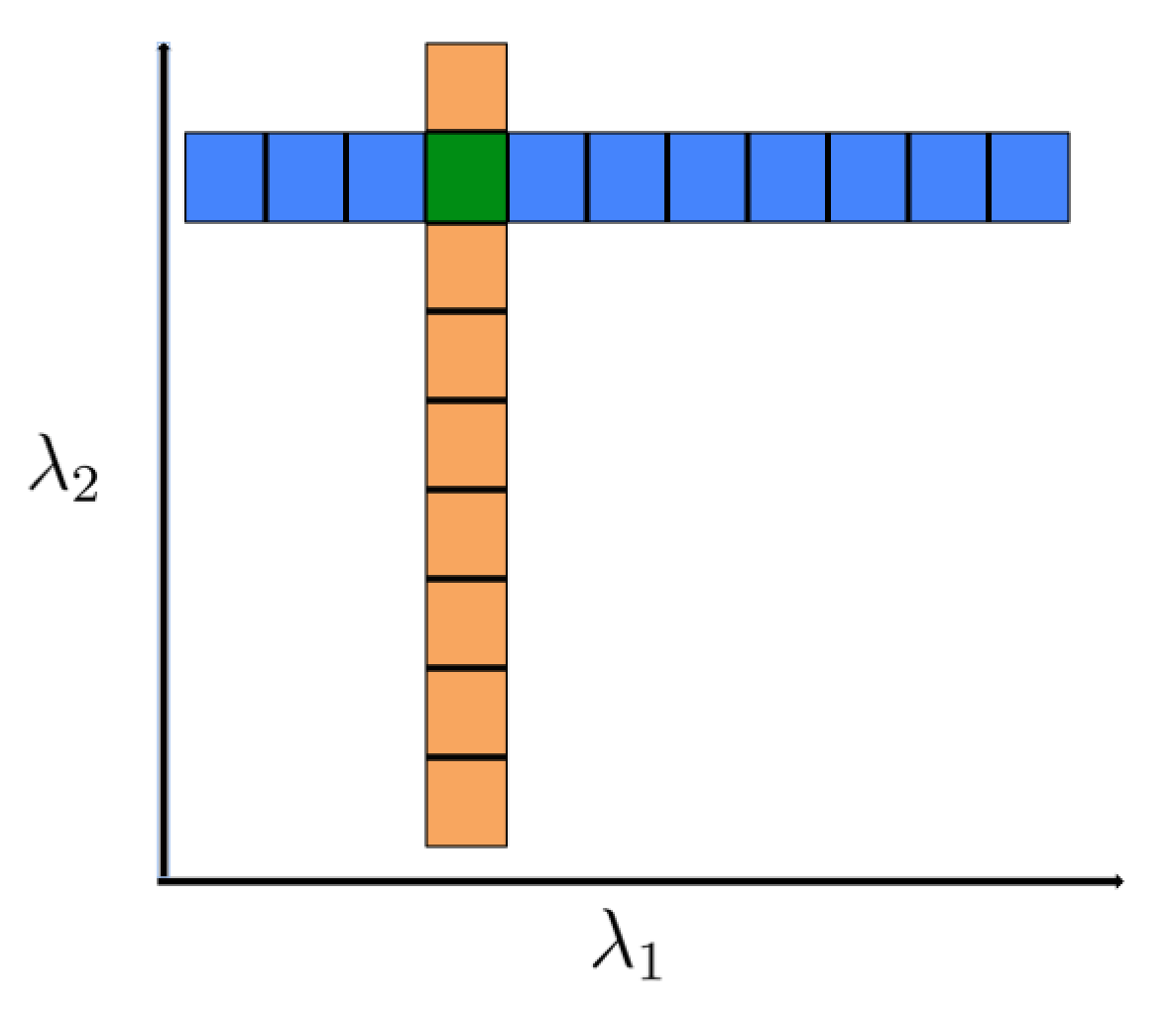}
	\caption{Graphical illustration of Greedy Local Continuation search. $\lambda_1$ and $\lambda_2$ are optimized alternatively with warm-starting strategy.}
	\label{fig:greedysearch}
\end{figure}

This warm start continuation strategy can also be applied on top of FAPGAR (Algorithm \ref{alg:FAPG}), when either $\lambda_1$ or $\lambda_2$ is fixed. Although it is possible to produce the solution path over the entire two dimensional grids generated by the Cartesian product $\lambda_1 \bigotimes \lambda_2$ with warm-starting strategy, the computation cost becomes large since $\card{(\lambda_1)} \times \card{(\lambda_2)}$ calls of Algorithm \ref{alg:FAPG} is required. Here $\card{(\lambda)}$ is the cardinality of grids in $\lambda$. Hence, we develop a greedy strategy with continuation method to avoid the computation over the entire search space. We propose a coordinate-wise search strategy. $\lambda_1$ and $\lambda_2$ is optimized alternatively using the continuation scheme while the other is being fixed. After a solution path over $\lambda_1$ is computed for a fixed $\lambda_2$, we pick the value of $\lambda_1$ yielding the best prediction on a separate validation set. After $\lambda_1$ is selected, the solution path for $\lambda_2$ is computed with the warm-start continuation strategy. The process continuous and search for $\lambda_1$  and $\lambda_2$ alternatively. Let $(\lambda^{(t)}_1,\lambda^{(t)}_2)$ denote the constants selected after $t$ coordinate-wise outer iterations. A cycle is a configuration $(\lambda^{(t)}_1,\lambda^{(t)}_2)$ such that $(\lambda^{(t)}_1,\lambda^{(t)}_2) = (\lambda^{(t')}_1,\lambda^{(t')}_2)$ for some $t'<t$. The search process stops when a cycle is detected. Figure \ref{fig:greedysearch} provides an illustration of the Greedy Local Continuation search.

\section{Case Studies}
\label{sec:case}
Experimental evaluations on the proposed low-rank and sparse decomposition-based latent effect model (\ref{eq:latent_lowrank_sparse}) and the solution algorithm FAPGAR are described in this section. Traffic accident records from The Statewide Integrated Traffic Records System (SWITRS) \citep{SWITRS} in California are used in the experiments. It has been reported that single-vehicle crashes account for nearly 30\% of all vehicle crashes in the United States in 2015 \citep{NHTSA}. The latent effect logit model (\ref{eq:latent_lowrank_sparse}) is evaluated on single-vehicle crash observations. The set of injury categories are $\{$\texttt{no injury or only complaint of pain, visible injury, severe injury, fatal}$\}$. Seventeen variables listed in Table \ref{tab:features} are used in the model. All variables are processed into binary vector $\mathbf{x}_n$'s using dummy coding, in which variable $x_{ni}$ is 0 if it belongs to the reference level in Table \ref{tab:features}, and 1 otherwise. We aim to study the computational aspect of (\ref{eq:latent_lowrank_sparse}) and the usage of this model for analyzing individual heterogeneity and quantifying the impact of different accident attributes. 

\begin{table*}[t!]
	\renewcommand{\arraystretch}{1.2}		
	\centering
	\begin{tabular}{@{}lllllllll@{}}
		\toprule
		feature ID &  0 & 1 & 2 & 3 & 4 & 5 & 6 & 7 \\ 
		\midrule	
		Description & \pbox{1.5cm}{victim\\gender} & \pbox{1cm}{victim\\age} & \pbox{1.5cm}{seatbelt\\used} & \pbox{1.4cm}{alcohol\\used} & \pbox{1.5cm}{cellphone\\used} & \pbox{1.5cm}{wrong \\side of \\road} & \pbox{1.5cm}{improper\\tuning} & \pbox{1cm}{over\\speed} \\ 
		\addlinespace[3pt]
		reference level & female & 25-64 & false & false & false & false & false & false \\  
		\hline 
	\end{tabular}
	\vspace{1pt}
	\begin{tabular}{@{}llllllllll@{}}
		\addlinespace[3pt]
		feature ID & 8 & 9 & 10 & 11 & 12 & 13 & 14 & 15 & 16\\ 
		\midrule	
		Description & \pbox{1.3cm}{weather} & \pbox{1.6cm}{light\\conditions} & \pbox{1.3cm}{wet\\road} & \pbox{1cm}{drug\\used} & \pbox{1cm}{vehicle\\age} &  \pbox{1cm}{AM\\peak}  & \pbox{1cm}{PM\\peak} & \pbox{1.5cm}{weekend} & \pbox{1cm}{season}    \\
		\addlinespace[3pt]
		reference level & clear & daylight & false & false & 0-10 & false & false & false & \pbox{1.2cm}{spring/\\summer} \\
		\addlinespace[1pt]
		\hline 
	\end{tabular}
	\caption{Accident-related features used in the experiment}
	\label{tab:features}
\end{table*}

\subsection{Computation efficiency}
\begin{figure}[ht!]
	\centering
	\includegraphics[width=0.5\textwidth]{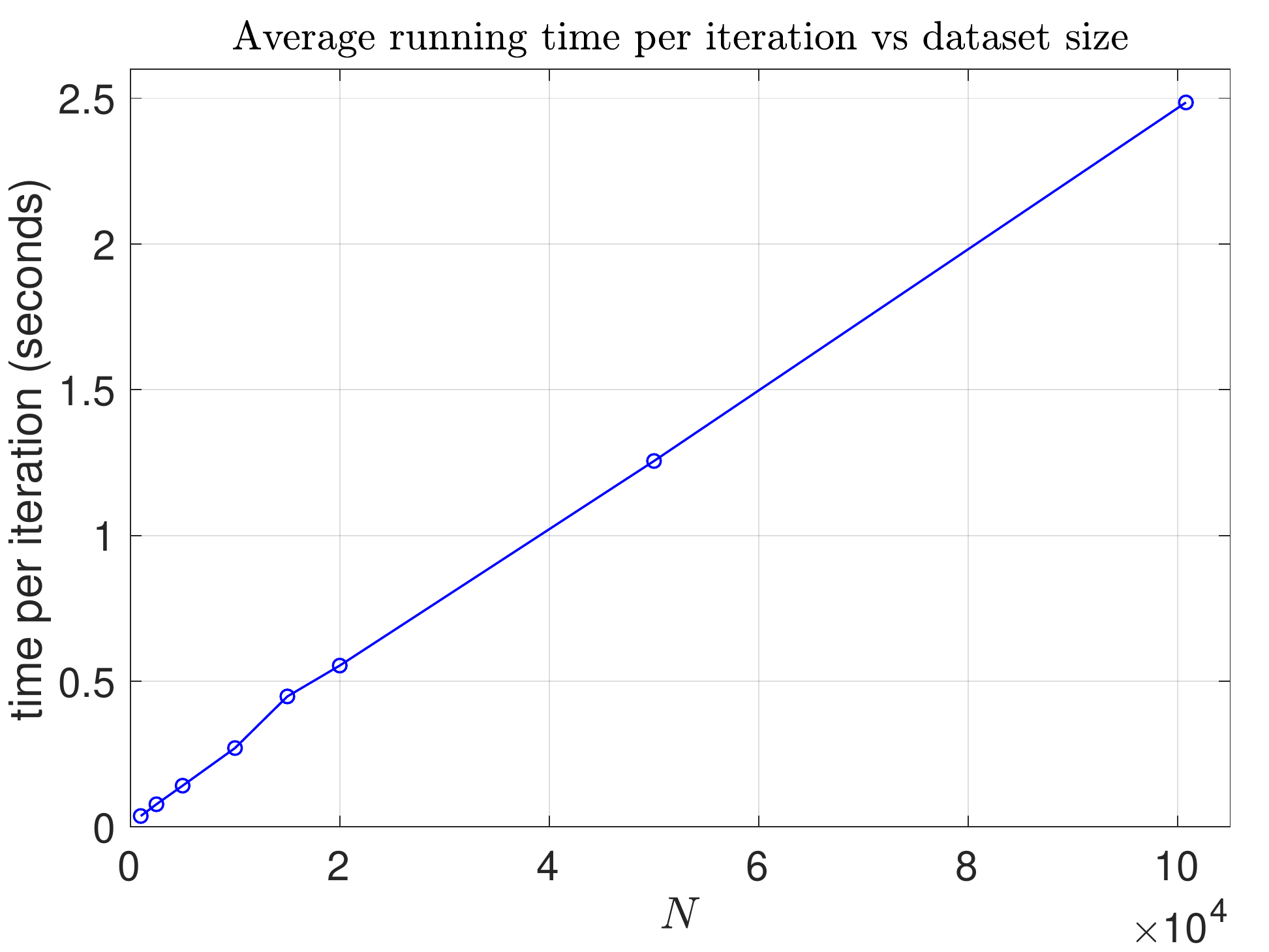}
	\caption{Time per iteration (seconds) of FAPGAR implemented in MATLAB.}
	\label{fig:FAPGAR_iterationtime}
\end{figure}

\begin{figure}[ht!]
	\centering
	\includegraphics[width=0.48\textwidth]{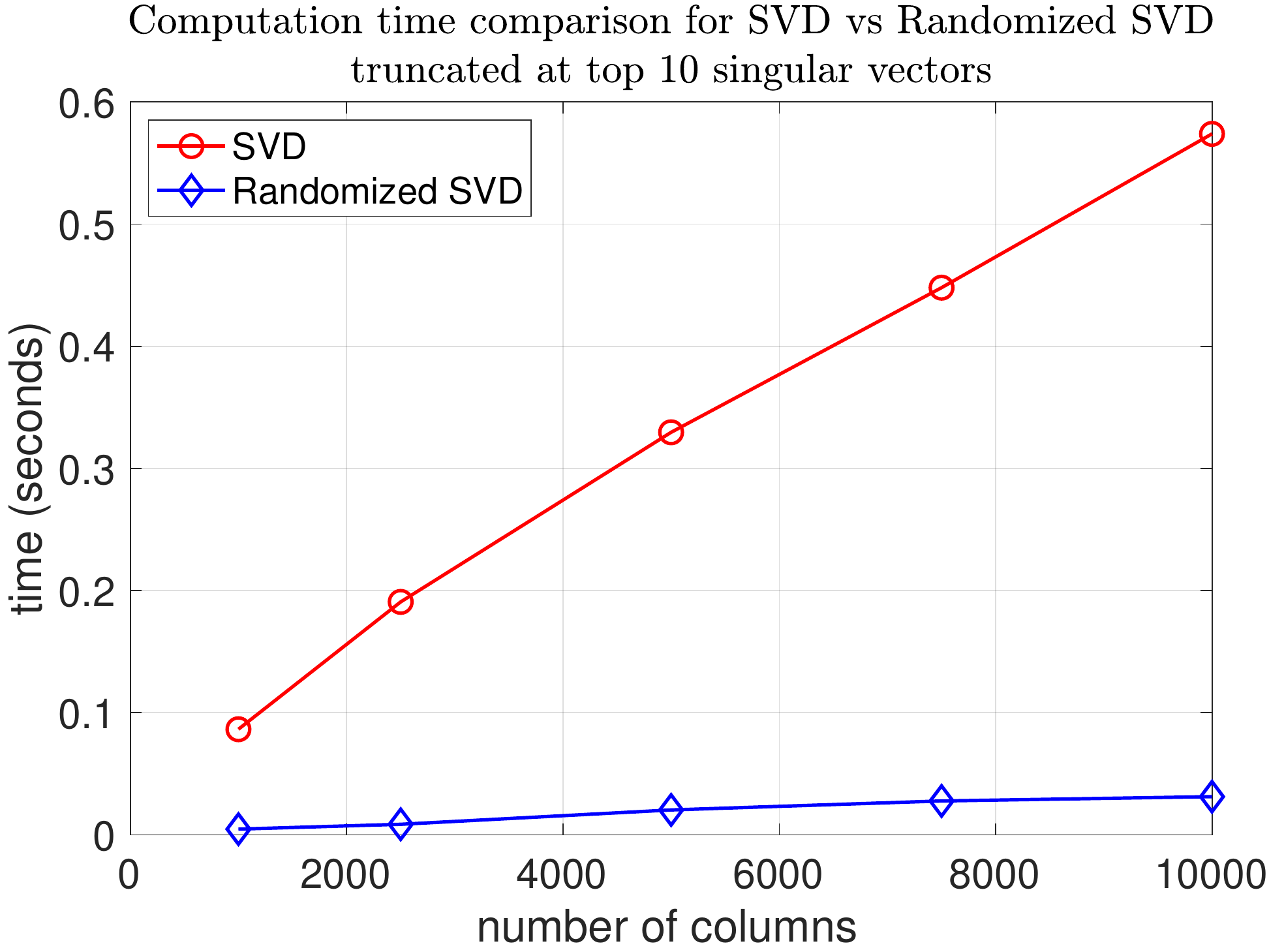}
	\includegraphics[width=0.48\textwidth]{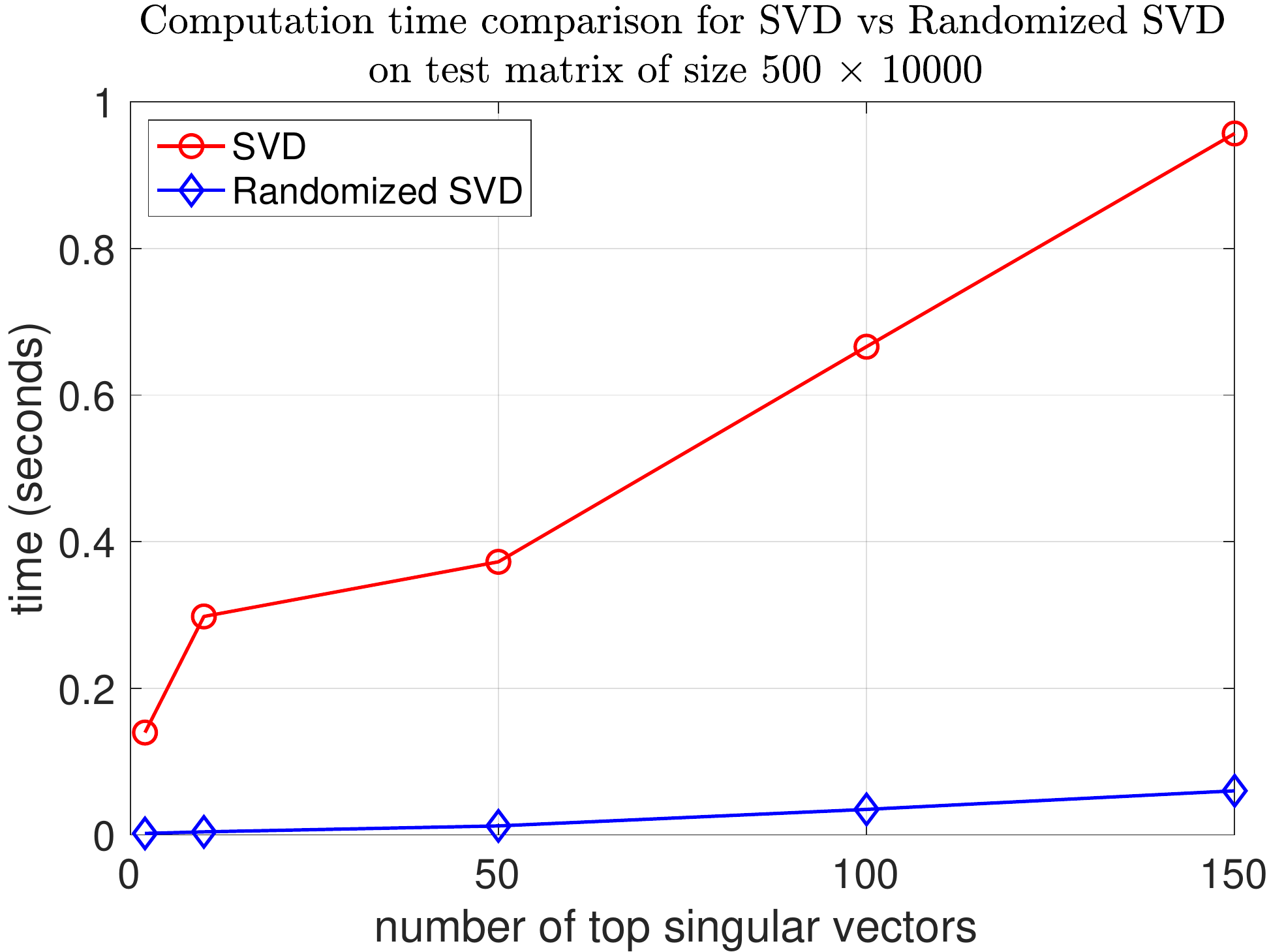}
	\caption{Computation time comparison, Randomized SVD vs SVD. Left: time against increasing number of columns in the input matrix with 500 rows. Right: time against increasing number of singular vectors required.}
	\label{fig:RSVDtime}
\end{figure}

\begin{figure}[ht!]
	\centering
	\includegraphics[width=0.5\textwidth]{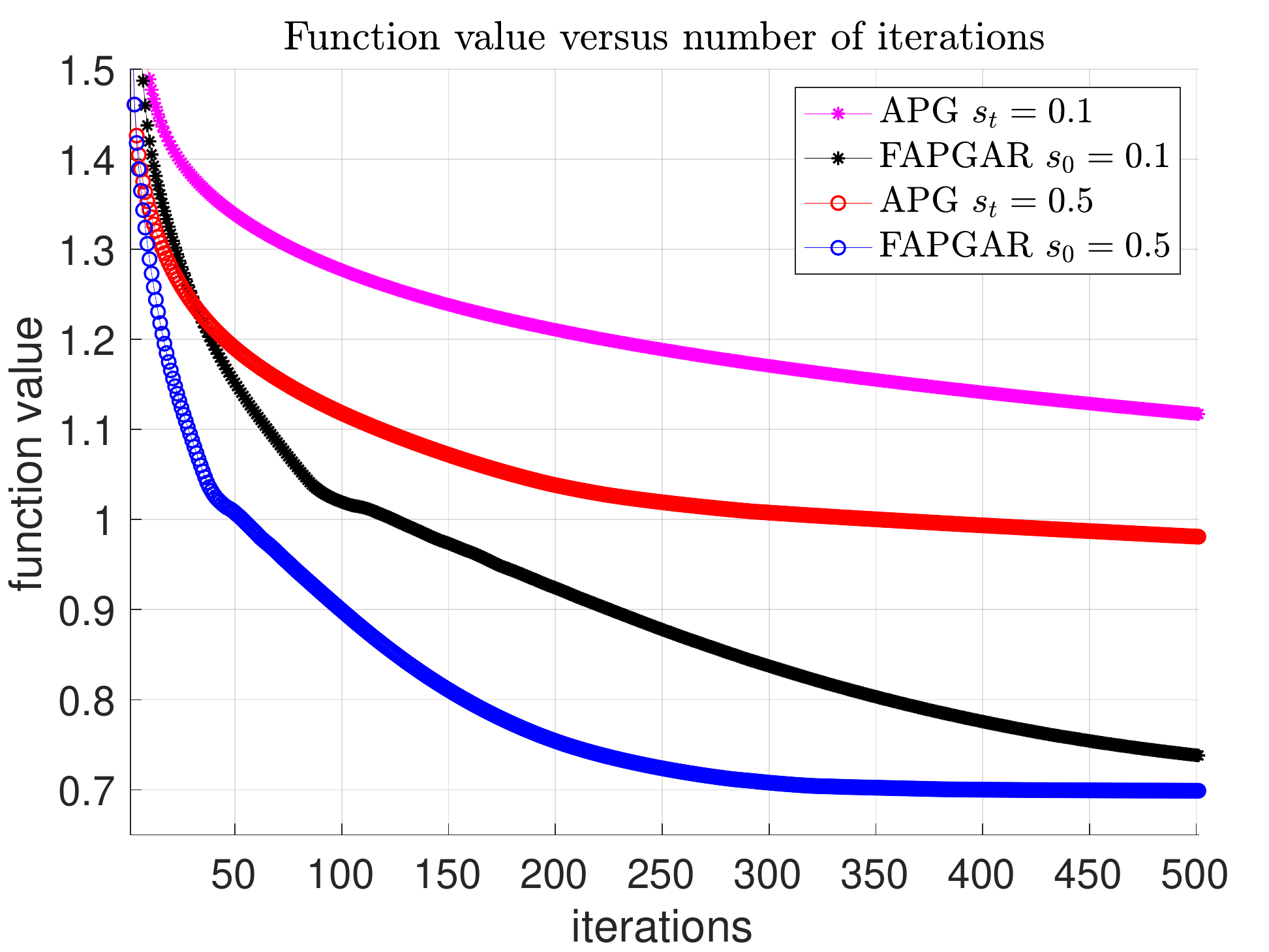}
	\caption{Objective value after each iteration, FAPGAR vs Proximal Gradient.}
	\label{fig:FAPGAR_loss}
\end{figure}

Computational efficiency of Algorithm \ref{alg:FAPG} is examined. We implement Algorithm \ref{alg:FAPG} in MATLAB and perform the experiments on a MacOS system with 2.4 GHz Intel Core i5 processor and 4 GB 1600 MHz DDR3 memory. Data from 2012-2013 are sub-sampled to create training sets with different sizes ($N$) in order to study the scalability of Algorithm \ref{alg:FAPG}. The average running time per iteration are shown in Figure \ref{fig:FAPGAR_iterationtime}. The computation time per iteration in FAPGAR scales linearly with the number of samples in the dataset. The FAPGAR algorithm is compared with the Proximal Gradient method with constant step-sizes. The objective value of (\ref{eq:latent_lowrank_sparse}) after each iteration is shown in Figure \ref{fig:FAPGAR_loss}.  Figure \ref{fig:FAPGAR_loss} clearly demonstrates the advantage of employing the adaptive acceleration techniques. The benefit of using Randomized SVD (Algorithm \ref{alg:randomizedSVD}) instead of deterministic SVD is illustrated in Figure \ref{fig:RSVDtime}. The computation time are measured on random input matrices. The left panel shows the time for performing randomized SVD implemented with Matlab and the deterministic partial SVD using Matlab command \texttt{svds} on matrices with varying number of columns. The right panel compares the randomized versus the deterministic approach at increasing number of required singular vectors. In both situations, randomized SVD offers more than a magnitude of speed-up. Note that there are approximation error introduced in the singular values computed via Randomzied SVD due to the random projection step in Algorithm \ref{alg:randomizedSVD}. However, the approximation error is relatively small as shown in Table \ref{tab:rsvderror}. 

To start the Local Greedy Continuation search for $\lambda_1$ and $\lambda_2$, we fist run group-$\ell_1$ regularized logistic regression over a range of $\lambda_1$s using software GLMNET \citep{friedman2007pathwise,friedman2010regularization}. Note that the result of group-$\ell_1$ regularized logistic regression is equivalent to our proposed model (\ref{eq:latent_lowrank_sparse}) with $\lambda_2$ set to a large value, which yields a matrix of zeros for the latent heterogeneity $\mathbf{\Upsilon}$. For each $\lambda_1$ on the solution path of group-$\ell_1$ regularized logistic regression, we classify the samples on a separate validation set by choosing the class with largest estimated probability as the output category for each validation sample. For each hyperparameter configuration, the \textbf{F-1 score} is computed:
\begin{equation}
\text{F-1 score} \eqdef \big(\frac{\text{recall}^{-1} + \text{precision}^{-1}}{2} \big)^{-1}.
\end{equation}

\begin{figure}[ht!]
	\centering
	\includegraphics[width=0.5\textwidth]{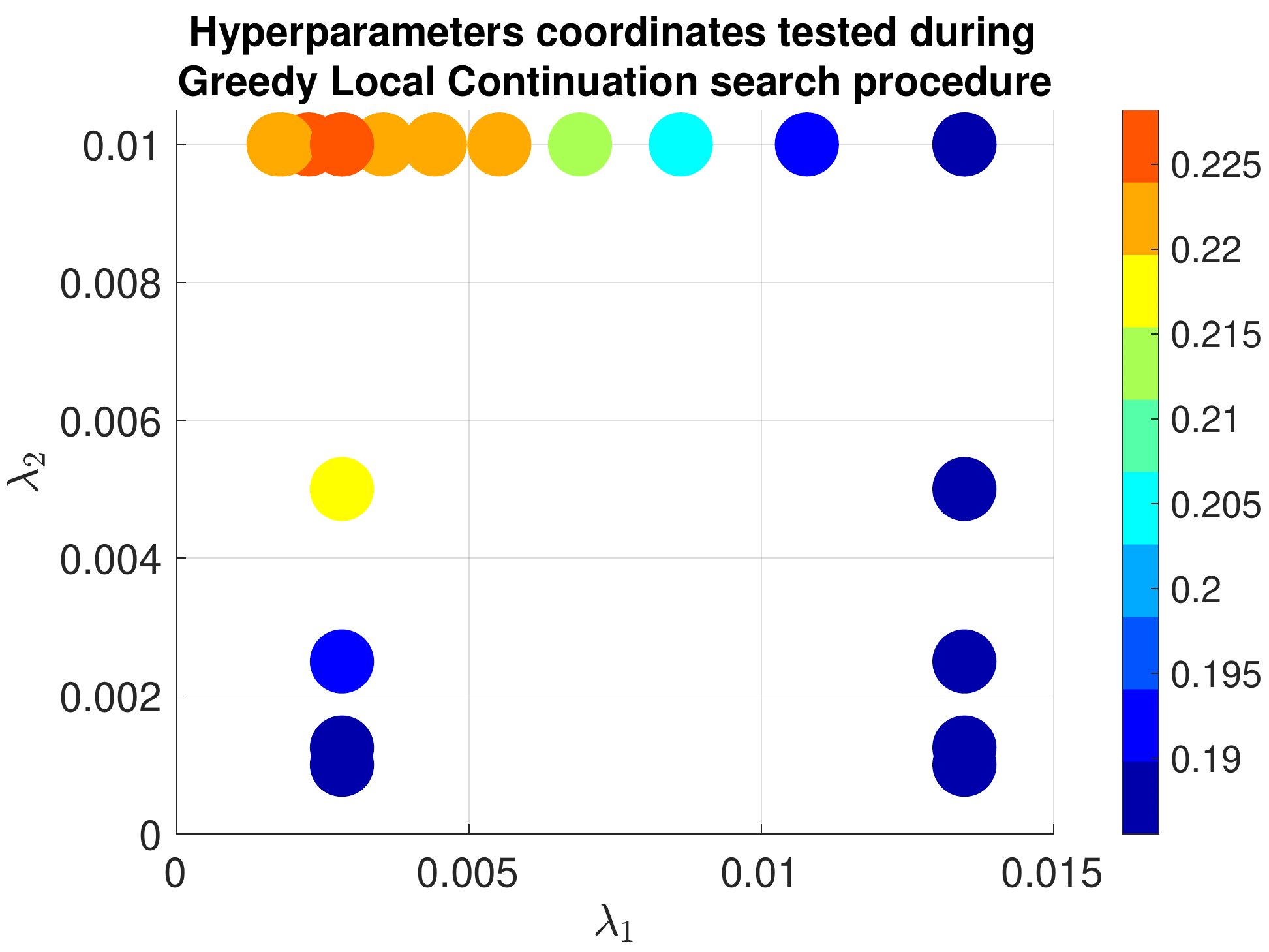}
	\caption{Greedy Local Continuation search on the traffic accident dataset, 2012-2013. A cycle is detected after three iterations.}
	\label{fig:greedysearch_MATLAB}
\end{figure}

The $\lambda_1$ value on the group-$\ell_1$ regularized logistic regression path with highest F-1 score is taken as initialization for the Local Greedy Continuation search. The progress of the greedy search process is visualized in Figure \ref{fig:greedysearch_MATLAB}. The search process terminated after three iterations due to the occurrence of a cycle. The selected $\lambda_1 = 0.0028, \lambda_2=0.01$.

\begin{table*}[ht!]
	\renewcommand{\arraystretch}{1.2}		
	\centering
	\begin{tabular}{llllll}
		\toprule
		   number of columns & 1000 & 2500 & 5000 & 7500 & 10000\\ 
		\midrule	
		   relative $\ell_2$ error & 0.0385  &  0.0337  & 0.0266  & 0.0230  & 0.0208 \\
	\end{tabular}
	\begin{tabular}{llllll}
	\toprule
	 top singular vectors & 2 & 10 & 50 & 100 & 150\\ 
	\midrule	
	relative $\ell_2$ error & 0.1120 & 0.0166 & 0.0216 & 0.0204 & 0.0157 \\
	\bottomrule		
	\end{tabular}
	\caption{Relative $\ell_2$ error in singular values computed with Randomized SVD.}
	\label{tab:rsvderror}
\end{table*}

\subsection{Accident factor analysis}

\begin{figure}[ht!]
	\centering
	\includegraphics[width=0.48\textwidth]{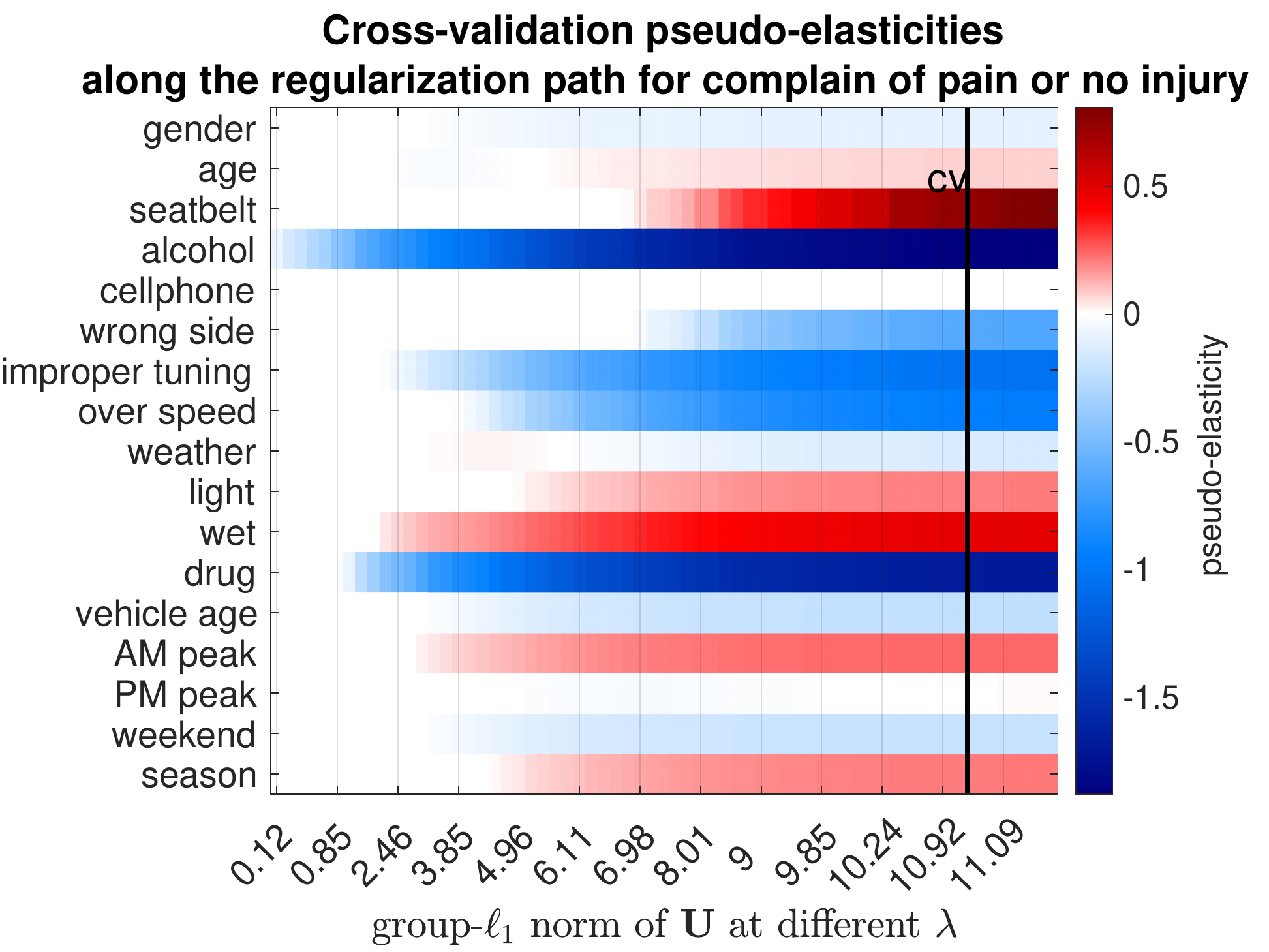}
	\includegraphics[width=0.48\textwidth]{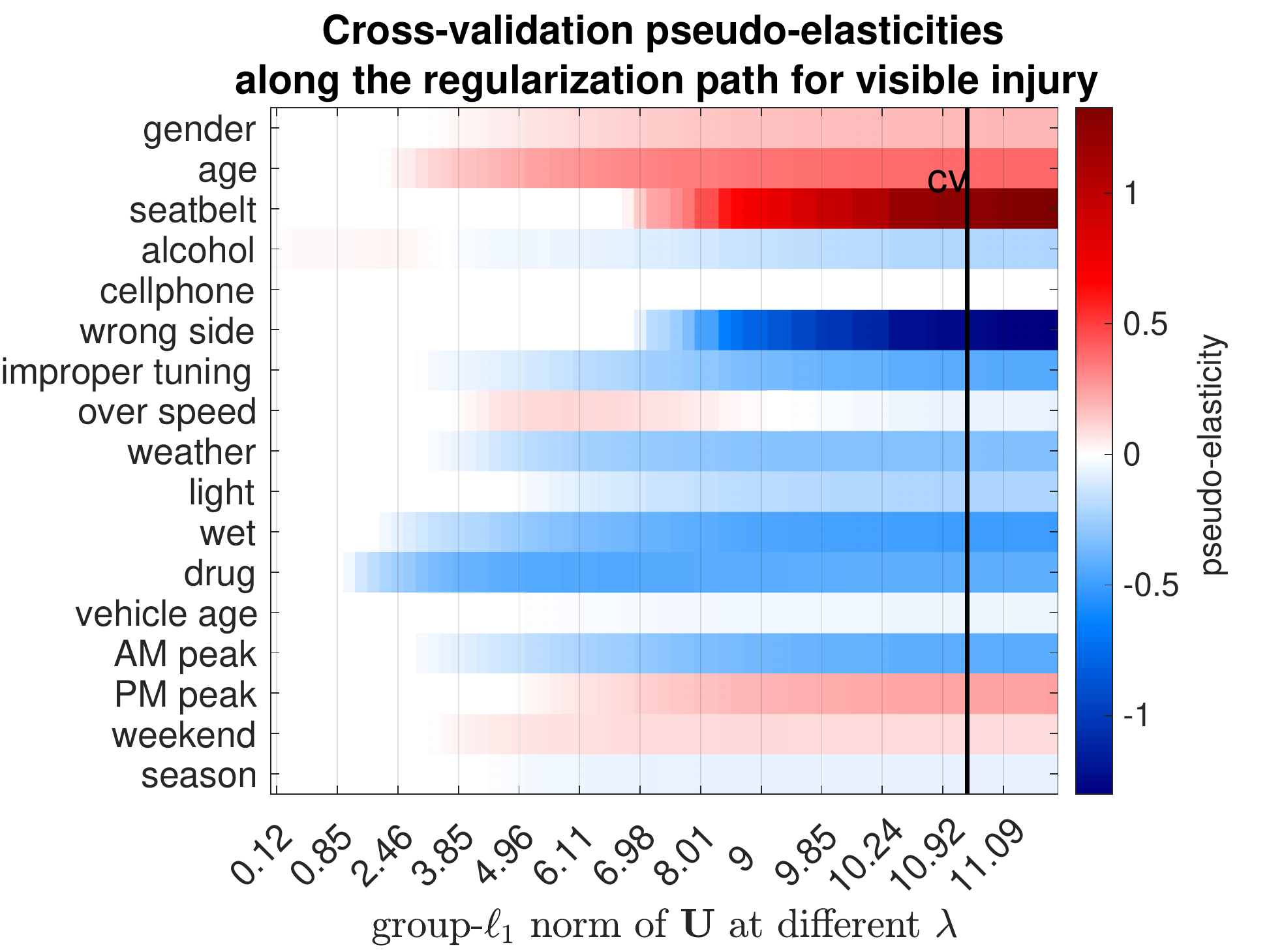}
	\includegraphics[width=0.48\textwidth]{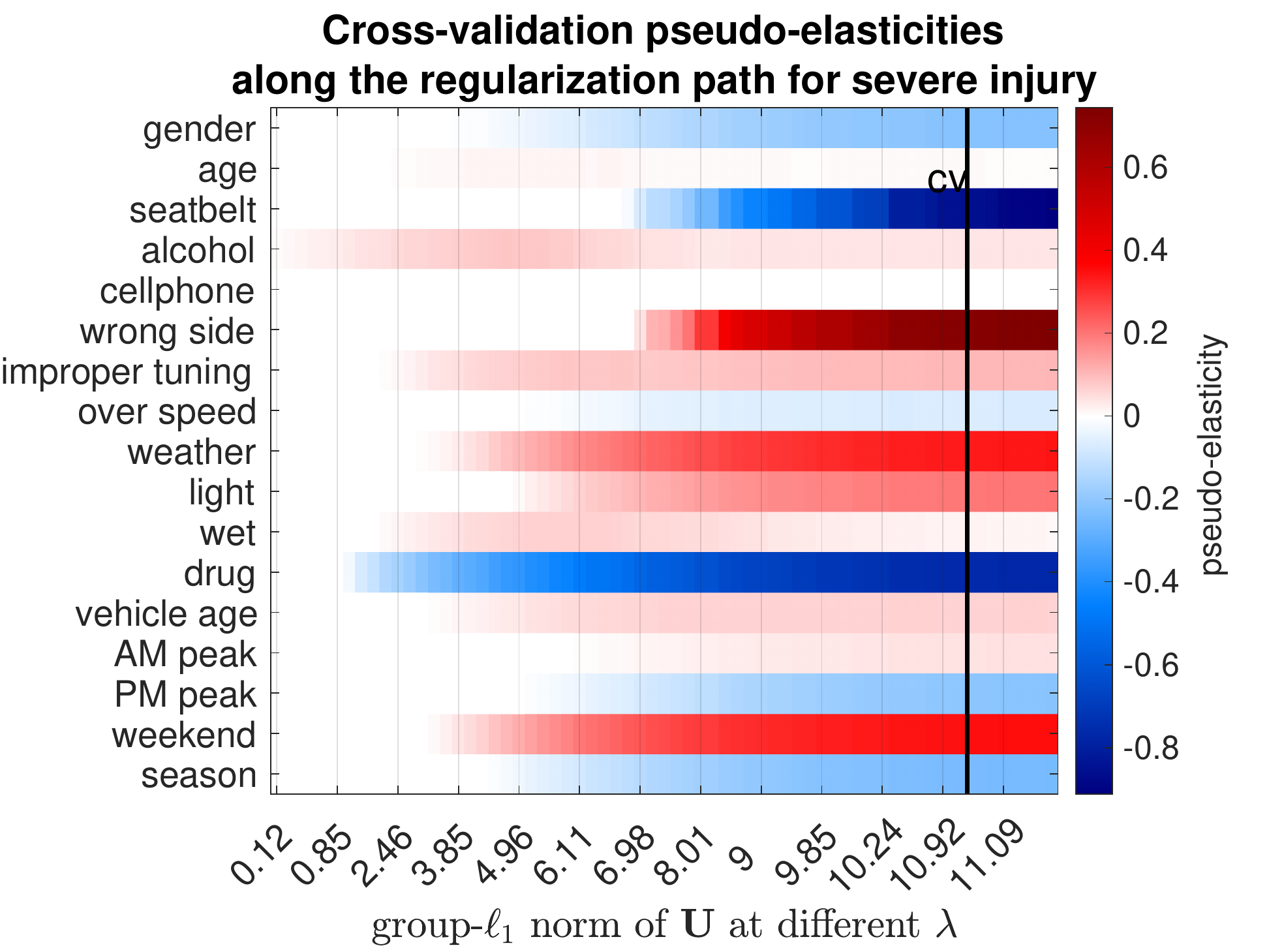}
	\includegraphics[width=0.48\textwidth]{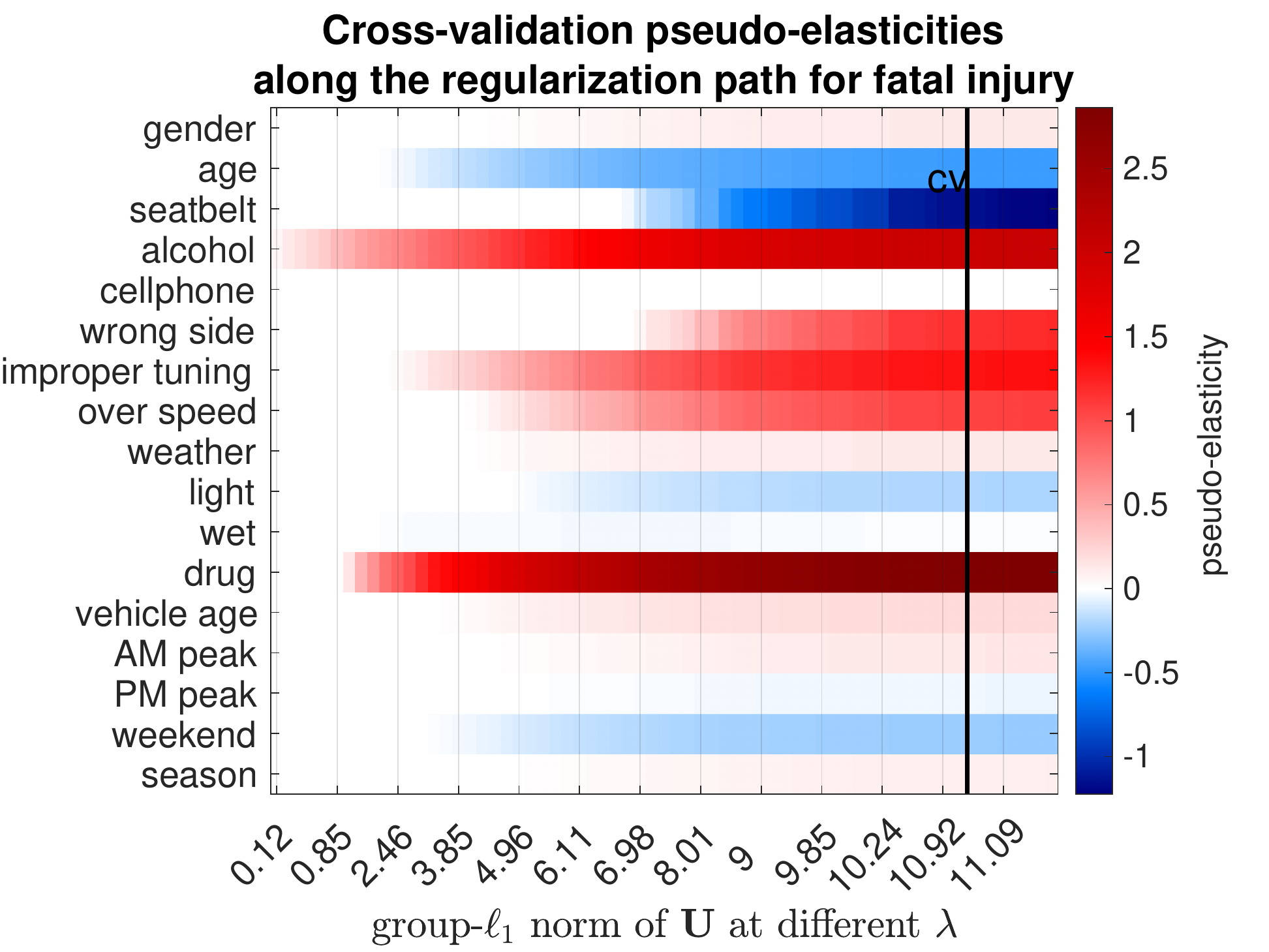}
	\caption{Cross-validation direct pseudo-elasticity over a path of $\lambda_1$. Vertical line marked with `CV' correspond to the $\lambda_1$ with best F-1 score from cross-validation.}
	\label{fig:cvelasticity}
\end{figure}

The latent effect model (\ref{eq:latent_lowrank_sparse}) is applied for analyzing traffic accident data in this section. The analysis is performed on $10000$ randomly sampled single-vehicle accidents from 2012-2013 in SWITRS. The average direct pseudo-elasticity is a metric for quantifying the influence of a factor \citep{shankar1996exploratory,shankar1998evaluating,carson2001effect,ulfarsson2004differences,anastasopoulos2009note,kim2010note,anastasopoulos2011empirical,morgan2011effects,kim2013driver,mannering2014analytic} (and references therein). 
Given a binary feature vector $\mathbf{x}$, the average direct pseudo-elasticity measures the change in the probability of suffering class $k$ injury outcome, when a feature $x_{i}$ switches from zero to one. The direct pseudo-elasticity for observation $n$ on class $k$ due to the change of $x_i$ can be computed by
\begin{equation}
\mathcal{E}^{(k)}_{ni} \eqdef \frac{P(t_n=k | \hat{\boldsymbol{\theta}}, \mathbf{x}_{n\backslash i},x_i = 1)-P(t_n=k | \hat{\boldsymbol{\theta}}, \mathbf{x}_{n\backslash i},x_i = 0)}{P(t_n=k | \hat{\boldsymbol{\theta}}, \mathbf{x}_{n\backslash i},x_i = 0)},
\end{equation}  
where $\hat{\boldsymbol{\theta}}$ denotes the estimated parameters for a model, $\mathbf{x}_{n\backslash i} \in \mathbb{R}^{p-1}$ is a sub-vector of features from $\mathbf{x}_n \in \mathbb{R}^p$ excluding the $i$-th variable. In transportation literature, the average of $\mathcal{E}^{(k)}_{ni}$ from the training set is calculated. Note that since the parameter $\hat{\boldsymbol{\theta}}$ is fitted from the training set, $\mathcal{E}^{(k)}_{ni}$ is in fact conditioned on the training set. The purpose of computing the average of $\mathcal{E}^{(k)}_{ni}$ is to achieve better estimation of the direct pseudo-elasticity in the population, i.e.,
\begin{equation}
\mathbb{E}_{n\sim \mathcal{D}} \Big( \mathbb{E}_{\mathcal{S} \sim \mathcal{D} }[\mathcal{E}^{(k)}_{ni} | \mathcal{S} ] \Big),
\end{equation}
where $\mathcal{D}$ denotes the population, and $\mathcal{S}$ denotes the training data randomly sampled from $\mathcal{D}$. Therefore, we propose an extension to the average pseudo-elasticity by incorporating the cross-validation procedure:
\begin{itemize}
	\item[1.] Shuffle the dataset randomly and separate it into $s$ folds. Let $\mathcal{S}_{\backslash i}$ denote the subset of data with the $i$-th fold removed, and $\mathcal{S}_{i}$ be the remaining data. 
	\item[2.] estimate a model on $\mathcal{S}_{\backslash i}$, and compute the average pseudo-elasticity on $\mathcal{S}_{i}$.
	\item[3.] compute the mean of average pseudo-elasticity estimated from $\mathcal{S}_{i},i=1,\cdots,s$.
\end{itemize} 
We name the estimated direct pseudo-elasticity from this procedure as the cross-validation direct pseudo-elasticity (CV-DPE). We compute CV-DPE of (\ref{eq:latent_lowrank_sparse}) over a path of $\lambda_1$ for the group-sparsity penalty constant, and fix $\lambda_2=0.01$ for the nuclear norm penalty constant obtained from the Greedy Local Continuation search. The results for each injury outcome category are displayed in Figure \ref{fig:cvelasticity}. Since the model parameters change when $\lambda_1$ varies, Figure \ref{fig:cvelasticity} provides a visual inspection of CV-DPE with different degrees of freedom in the model. From the CV-DPE analysis, alcohol increases the probability of suffering severe and fatal injuries. Meanwhile, the probability of experiencing only visible injuries and the chance of free from injuries in an accident is greatly reduce. In addition, alcohol is the first factor selected into the model, when $\lambda_1$ decreases. This is an indication about the importance of alcohol factor in predicting the accident results. On the contrary, wearing seatbelt improves the chance of no injury or suffering only visible injuries from accidents by more than $50\%$, reducing the odds of suffer severe injuries or fatality by more than $50\%$ according to the CV-DPE result from the estimated model with best F-1 score. Consumption of drugs increases the likeliness of fatality by more than $200\%$ according to the model selected by the cross-validation, while reducing the chance of all other injury categories. Factors related to violation of traffic laws, i.e., over-speeding, improper tuning, wrong side of road, all lead to greater chance of fatality, but relatively less impactful compared to the influence of drug and alcohol. In addition, the Local Greedy Continuation found a $\lambda_1$ that leads to a vector of zeros for the homogeneous effect parameter $\mathbf{U}_i$ corresponds to cell-phone usage. This can be interpreted as the exclusion of cell-phone influence as a common effect. This observation may be unintuitive at the first sight. After inspecting the dataset, only less than $1\%$ accident samples reported cell-phone as involved. This can be a result of under-reporting --- understandably, people would not admit using a cell-phone while driving unless being caught. Hence, it is not surprise to see the model selects cell-phone usage only as a heterogeneous effect variable.

\begin{figure}[ht!]
	\centering
	\includegraphics[width=0.5\textwidth]{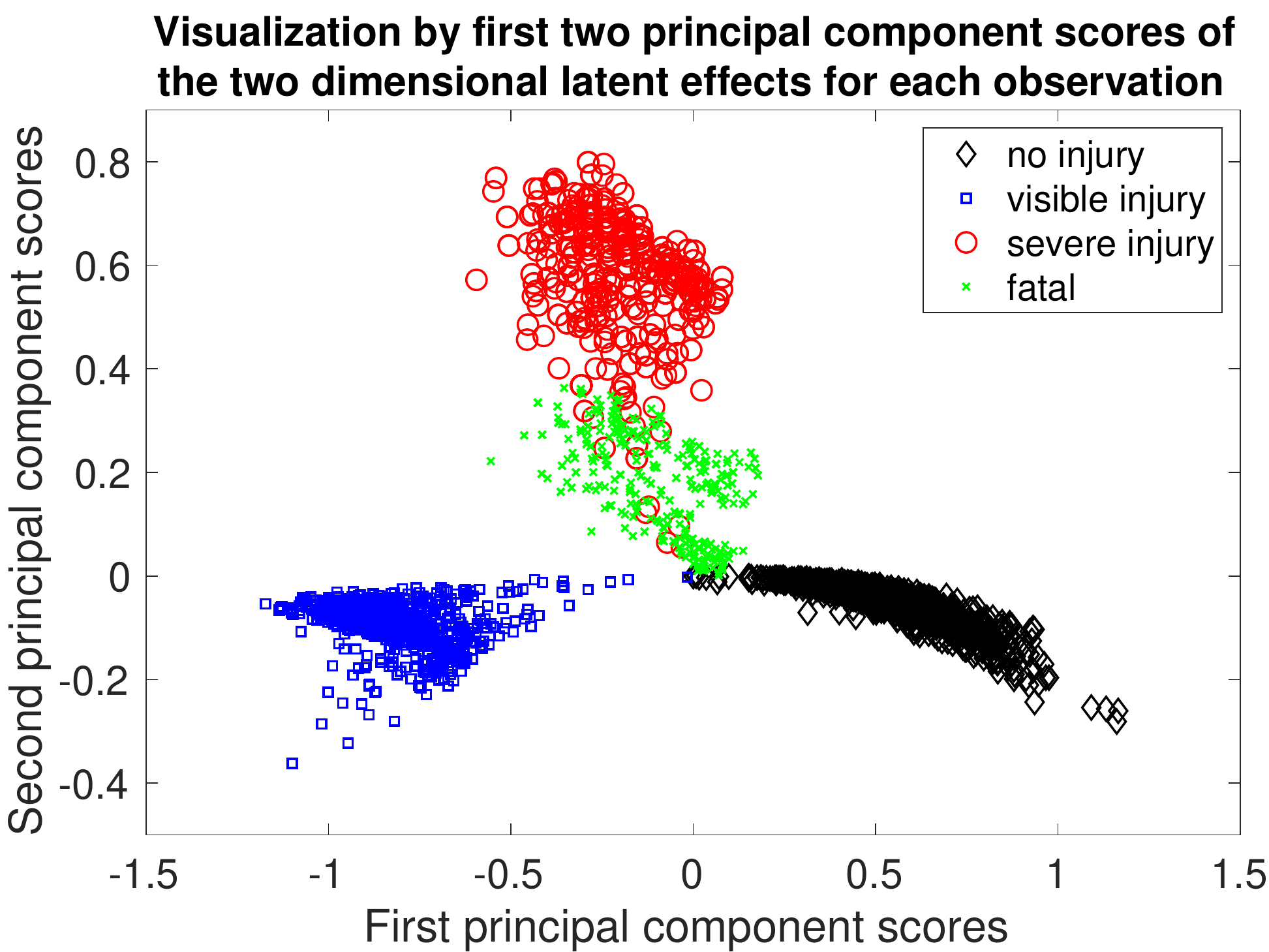}
	\caption{Latent heterogeneity $\boldsymbol{\upsilon}_n$ inferred from the 2012-2013 sample. $\rank(\mathbf{\Upsilon})=2$. Each point is represented by the first two principal component scores.}
	\label{fig:latent_pca}
\end{figure}

\subsection{Visualization of latent individual heterogeneity}
The latent effect model (\ref{eq:latent_lowrank_sparse}) also has the advantages of providing insights into the individual heterogeneity. The rank of $\mathbf{\Upsilon}$ deceases as $\lambda_2$ increases. Hence, (\ref{eq:latent_lowrank_sparse}) can be used as a dimensional reduction tool. The resulting $\mathbf{\Upsilon}$ learned from model (\ref{eq:latent_lowrank_sparse}) has rank two at the $\lambda_2$ selected from the Local Greedy Continuation search process. Hence, each $\boldsymbol{\upsilon}$ can be visualized by a two dimensional vector, represented by their principal component scores. The first two principal component scores of the latent heterogeneity $\{\boldsymbol{\upsilon}_n\}_{n=1}^N$ inferred from each training point are shown in Figure \ref{fig:latent_pca}. There are four well identified clusters, each cluster is dominated by samples from one injury category. This is an indication of the clustering effect in the individual heterogeneity. Moreover, the cluster with majority of severe injury accidents and the cluster comprises mostly fatality cases are overlapped with each other.

\section{Conclusion}
\label{sec:conclusion}
We present a latent effect logistic model based on sparse and low-rank decomposition between the homogeneous effect and heterogeneous effect of observations. The formulation has the advantages of preserving model convexity while capturing the latent individual heterogeneity. The optimization problem from the model is solved by a Fast Accelerated Proximal Gradient with Adaptive Restarting algorithm. A Greedy Local Continuation search process is developed to enable efficient exploration of model hyperparameters. We demonstrate that low-rankness is a result of different data-generating processes, and validate through experiments clustering gave rise to the low-rankness of latent heterogeneity in accident observations. The usefulness of the model is demonstrated by analyzing traffic accident factors. From the analysis, drug and alcohol are found to be the factors with the largest impact on the probability of suffering fatality in an accident, whereas the usage of seatbelt greatly improves the chance of avoiding injuries.

\bibliographystyle{plainnat}
\bibliography{splrref}
\end{document}